\documentclass[10pt,twocolumn,letterpaper]{article}

\usepackage{cvpr}              

\makeatletter
\@namedef{ver@everyshi.sty}{}
\makeatother
\usepackage{tikz}

\usepackage{graphicx}
\usepackage{amsmath}
\usepackage{amssymb}
\usepackage{booktabs}
\usepackage{multirow}
\usepackage[utf8]{inputenc}

\usepackage[pagebackref,breaklinks,colorlinks]{hyperref}

\usepackage[capitalize]{cleveref}
\crefname{section}{Sec.}{Secs.}
\Crefname{section}{Section}{Sections}
\Crefname{table}{Table}{Tables}
\crefname{table}{Tab.}{Tabs.}

\newcommand{\keypoint}[1]{\vspace{0.1cm}\noindent\textbf{#1}\quad}
\newcommand{\cut}[1]{}

\newcommand{\ourName}{P>M>F}
\newcommand{\ourNameLong}{Pre-training $\rightarrow$ Meta-training $\rightarrow$ Fine-tuning~}

\newcommand\circled[1]{%
  \tikz[baseline=(X.base)] 
    \node (X) [draw, shape=circle, inner sep=-1, fill=white, text=black] {\strut \footnotesize #1};%
}
\usepackage{soul}
\usepackage{tabularx}

\def\cvprPaperID{9441} 
\def\confName{CVPR}
\def\confYear{2022}

\usepackage{overpic}
\usepackage{enumitem} 
\usepackage{overpic} 
\usepackage{color}

\usepackage[utf8]{inputenc} 
\usepackage[T1]{fontenc}    
\usepackage{url}            
\usepackage{booktabs}       
\usepackage{amsfonts}       
\usepackage{nicefrac}       
\usepackage{microtype}      
\usepackage{xcolor}         

\usepackage{graphicx}
\usepackage{amsmath}
\usepackage{multirow}
\usepackage{caption}
\usepackage{pifont}
\usepackage{bbm}

\usepackage{times}
\usepackage{amssymb}
\usepackage{comment}
\usepackage{colortbl}
\usepackage{enumitem}
\usepackage{makecell}
\usepackage{arydshln}

\usepackage{algorithm}
\usepackage{bbding}
\usepackage{listings}

\definecolor{turquoise}{cmyk}{0.65,0,0.1,0.3}
\definecolor{purple}{rgb}{0.65,0,0.65}
\definecolor{dark_green}{rgb}{0, 0.5, 0}
\definecolor{orange}{rgb}{0.8, 0.6, 0.2}
\definecolor{red}{rgb}{0.8, 0.2, 0.2}
\definecolor{darkred}{rgb}{0.6, 0.1, 0.05}
\definecolor{blueish}{rgb}{0.0, 0.3, .6}
\definecolor{light_gray}{rgb}{0.7, 0.7, .7}
\definecolor{pink}{rgb}{1, 0, 1}
\definecolor{greyblue}{rgb}{0.25, 0.25, 1}

\newcommand{\Table}[1]{Table~\ref{tab:#1}}

\usepackage{blindtext}

\renewcommand{\paragraph}[1]{\vspace{1em}\noindent\textbf{#1}.}

\begin{document}

\title{Pushing the Limits of Simple Pipelines for Few-Shot Learning:\\
External Data and Fine-Tuning Make a Difference}
\author{
Shell Xu Hu$^1$
\and 
Da Li$^{1}$\thanks{Equal contributions.} 
\and 
Jan St\"uhmer$^{1*}$ 
\and 
Minyoung Kim$^{1*}$ 
\and 
Timothy M. Hospedales$^{1,2}$
\and
$^1$Samsung AI Center Cambridge \qquad $^2$University of Edinburgh\\
{\tt\small \{shell.hu, da.li1, jan.stuhmer, k.minyoung, t.hospedales\}@samsung.com}
}

\maketitle
\begin{abstract}
Few-shot learning (FSL) is an important and topical problem in computer vision that has motivated extensive research into numerous methods spanning from sophisticated meta-learning methods to simple transfer learning baselines. We seek to push the limits of a simple-but-effective pipeline for more realistic and practical settings of few-shot image classification. To this end, we explore few-shot learning from the perspective of neural network architecture, as well as a three stage pipeline of network updates under different data supplies, where unsupervised external data is considered for pre-training, base categories are used to simulate few-shot tasks for meta-training, and the scarcely labelled data of an noval task is taken for fine-tuning.
We investigate questions such as: \circled{1} How pre-training on external data benefits FSL? \circled{2} How state-of-the-art transformer architectures can be exploited?
and \circled{3} How fine-tuning mitigates domain shift? Ultimately, we show that a simple transformer-based pipeline yields surprisingly good performance on standard benchmarks such as Mini-ImageNet, CIFAR-FS, CDFSL and Meta-Dataset.
Our code and demo are available at \url{https://hushell.github.io/pmf}.

\end{abstract}

\section{Introduction}
\label{sec:intro}

\begin{figure}[t]
\begin{center}
\includegraphics[width=\linewidth]{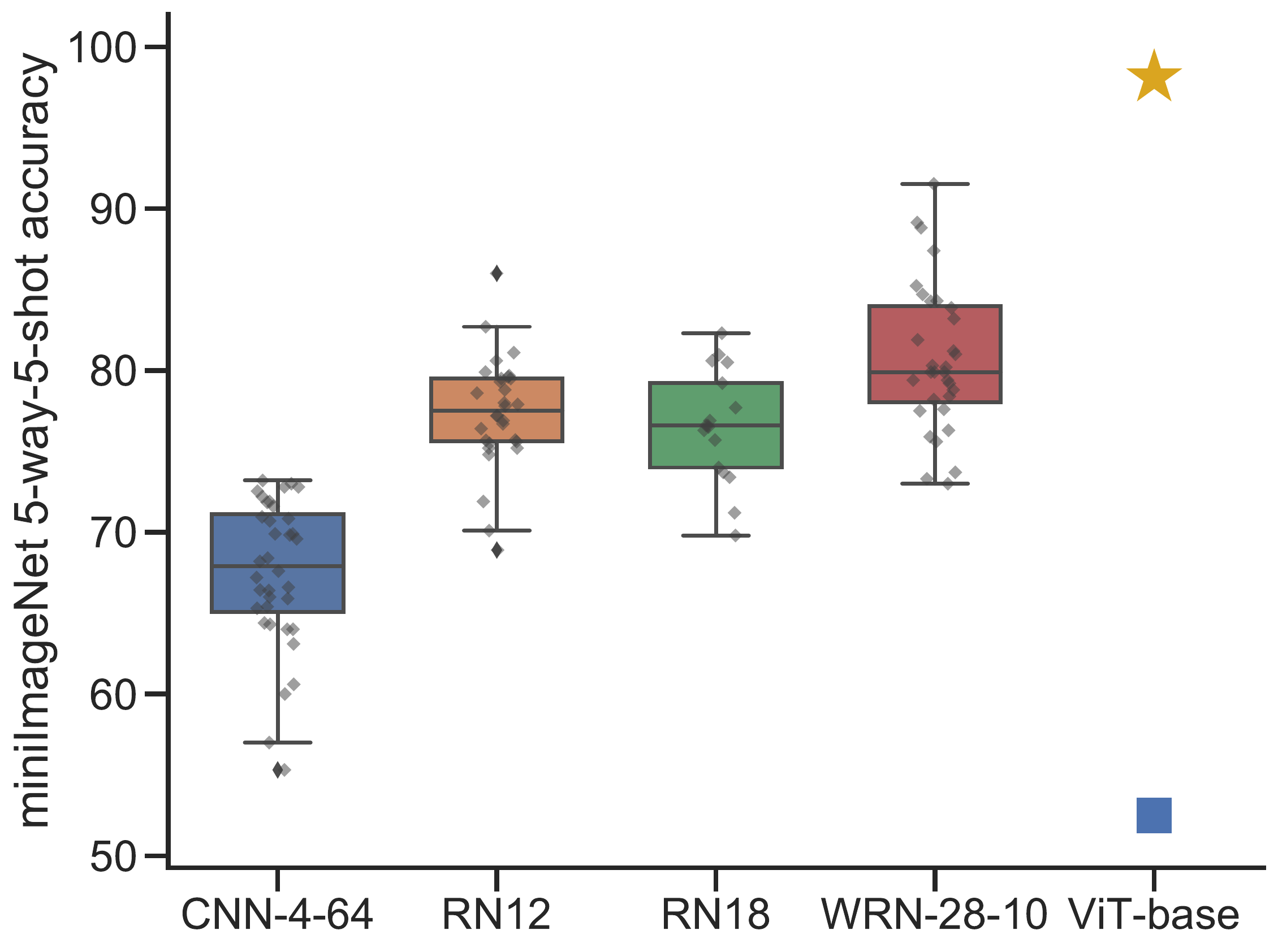}
\end{center}
\caption{
\textbf{How does pre-training and architecture affect few-shot learning?} 
Learning from a few shots can be achieved by a) meta-learning \cite{vinyals2016matching,zhang2021metaQDA} and b) transfer learning from self-supervised foundation models pre-trained on large-scale external data \cite{chen2020bigSimCLRV2,radford2021learning}. While the majority of FSL community focuses on the former, we show that the latter can be more effective because it enables the use of stronger architectures such as vision transformer (ViT) \cite{dosovitskiy2021imageTransformerVIT} -- and can be combined with simple meta-learners such as ProtoNet. The figure shows results aggregated from dozens of studies from the past 5 years of FSL research and the result of ProtoNet + ViT backbone + contrastive language-image pretraining (CLIP) \cite{radford2021learning} (yellow star). To emphasize the importance of pre-training, ProtoNet + randomly initialized ViT (blue square) is also compared.  
}
\label{fig:teaser}
\end{figure} 

Mainstream supervised deep learning achieves excellent results in applications where huge annotated datasets are available. However, this assumption is not met in many applications where data (e.g., rare categories), or the cost of human annotation are prohibitive bottlenecks. This has motivated a large and growing set of research in few-shot learning (FSL), which aims to emulate the human ability to learn new concepts from few training examples. The FSL challenge has proven fertile ground for developing and testing a vast array of sophisticated research ideas   spanning metric learning \cite{snell2017prototypical,sung2018relationNet}, gradient-based meta-learning \cite{finn2017model}, program induction \cite{lake2015ppi}, differentiable optimization layers \cite{lee2019meta}, hypernetworks \cite{bertinetto2016feedForwardOneShot}, neural optimizers \cite{ravi2017fewShotMeta}, transductive label propagation \cite{ren2018metaSSL}, neural loss learning \cite{baik2021metalFSL}, Bayesian neural priors \cite{zhang2021metaQDA} and more  \cite{wang2020fslSurvey}. But how much practical progress have we made based on all these technical advances? 

A few studies \cite{chen2019closerLook,chen2021metaBaseline,dhillon2020baselineFSL,mangla2020charting,wang2019simpleshot,tian2020rethinking} have investigated whether  simpler baselines can offer comparable performance to sophisticated state of the art few-shot learners. While there is no conclusive answer, due to on-going developments in both sophisticated learners \cite{zhang2021metaQDA} and simple baselines, there is a trend that simple approaches often perform surprisingly well compared to sophisticated counterparts. Their simplicity and efficacy leads these simple methods to be taken up in many practical applications of few-shot learning from medical data analysis \cite{bontonou2020fsBrain} to electronic engineering \cite{kazemi2021}. 

We follow this line of enquiry, but go further in investigating previously under-studied factors that influence the performance of simple few-shot pipelines. In particular we start with a ProtoNet \cite{snell2017prototypical} few-shot learner, and investigate three practically important design choices: pre-training data, neural network architecture, and meta-test time fine-tuning. %

\noindent\textbf{Source data}\quad
While FSL addresses the small data regime, in reality FSL research is almost always about algorithms to transfer knowledge from large scale source tasks (aka meta-train) to small scale target tasks (aka meta-test). Existing literature almost always controls the source data, in order to carefully compare the impact of different knowledge transfer mechanisms of interest from hyper-networks \cite{bertinetto2016feedForwardOneShot} to gradient-based meta-learners \cite{finn2017model}. While this is helpful to drive research on sophisticated \emph{algorithms}, it does not answer the question of how choice of source \emph{data} impacts performance? This question has been studied in other areas of vision and pattern recognition \cite{goyal2019scaling,sun2017unreasonable,bommasani2021foundation}, but not for FSL. This is unhelpful for consumers of computer vision FSL research, who would be  interested to know how much a simple change of source data can improve their applications? Especially since freely available large datasets already exist \cite{deng2009imagenet,thomee2016yfc100m}, and {exploiting more external source data is easier in practice than implementing sophisticated state-of-the-art meta-learners}. To this end we investigate the impact of unsupervised pre-training on external data -- a workflow recently termed as exploiting a \emph{foundation model} \cite{bommasani2021foundation} -- on FSL tasks. This small change has substantial impact compared to 5 years of FSL research (Figure~\ref{fig:teaser}). %
Although this may violate definitions of the FSL problem that strictly prescribe the source set, the efficacy of the approach may prompt reflection on whether this is the best problem definition to focus on.

\keypoint{Neural architecture} 
Similarly to the situation with source data, FSL studies often control neural architecture to a handful of small networks such as CNN-4-64 and ResNet-12. This is partly to enable fair comparison of FSL algorithms, but this particular suite of networks is also a consequence of the small size of the source datasets used for training in common benchmarks such as miniImageNet. Thus the architectures commonly studied in FSL are somewhat out-of-date with regard to state-of-the-art computer vision. We therefore ask to what extent state-of-the-art architectures such as vision transformers \cite{dosovitskiy2021imageTransformerVIT} can benefit few-shot performance, especially in conjunction with larger pre-training datasets? 

\keypoint{Fine-tuning} 
The many studies in the FSL literature are somewhat divided in whether they advocate \cite{finn2017model,ravi2017fewShotMeta,triantafillou2019meta} some kind of fine-tuning during model deployment (aka meta-test) for individual tasks, or whether a fixed feature representation should be sufficient \cite{wang2019simpleshot,snell2017prototypical,lee2019meta}. We also investigate this issue, 
and suggest that 
\emph{fine-tuning is %
 necessary for deploying foundation models to out-of-distribution tasks}. 
We also introduce an algorithmic improvement to fine-tuning by automating the learning rate selection via validation, which leads to 
a more performant pipeline for cross-domain FSL.

\begin{figure}
\begin{center}
\includegraphics[width=1.0\columnwidth]{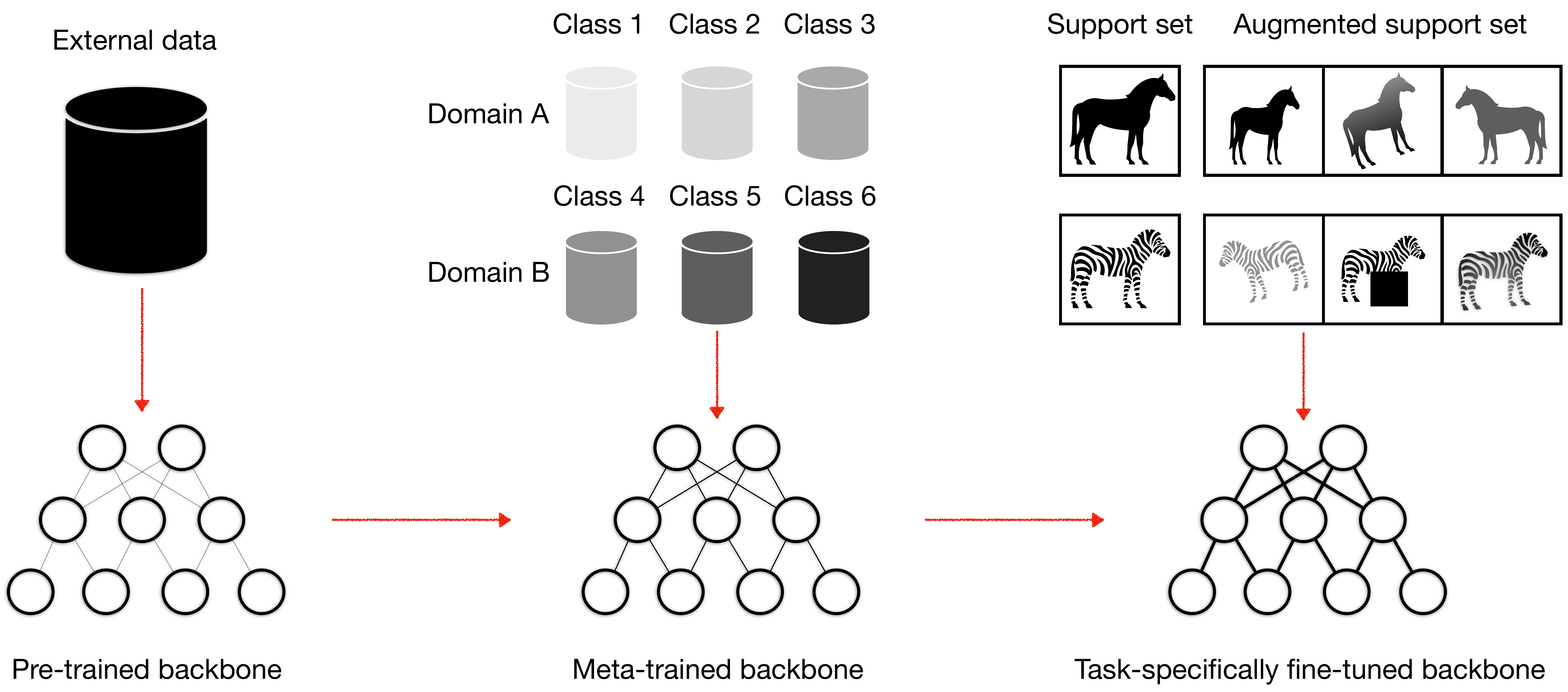}
\end{center}
\caption{
    \textbf{Overview -- } A schematic of the simple-but-effective pipeline that we consider: \ourNameLong (\ourName). Following the red arrows, the pipeline turns a class-agnostic feature backbone into a generic feature backbone and ultimately a task-specific feature backbone. 
}
\label{fig:overview}
\end{figure}

In summary, we advance few-shot learning by studying design choices of a simple pipeline \cite{snell2017prototypical}  (Figure~\ref{fig:overview}), rather than developing new algorithms. We answer questions including: \emph{How does pre-training impact FSL?} \emph{Can recent transformer architectures be adapted to FSL?} and \emph{How to best exploit fine-tuning?} 
Based on this analysis we demonstrate a new baseline for FSL that  surpasses state-of-the-art performance, while being simple and easy to implement. %

\section{Related Work}\label{sec:related}

\keypoint{Few-shot learning}
Few-shot learning is now a deep and widely studied area too large to review in detail here, and we refer to relevant surveys for an overview \cite{wang2020fslSurvey,hospedales2020meta}. A key point is that, despite the name, almost all FSL methods provide algorithms for transferring knowledge from a large set of source data, to a set of sparsely annotated target categories of interest. Much activity in the field falls under the umbrella of meta-learning \cite{hospedales2020meta}, which aims to construct a data-efficient learner from the source (aka meta-train) dataset by simulating few-shot learning problems, and then deploy the customized learner on the target  (aka meta-test)  set. The resulting learner may take the form of an initialization \cite{finn2017model}, learned metric \cite{snell2017prototypical}, Bayesian prior \cite{zhang2021metaQDA}, or optimizer \cite{ravi2017fewShotMeta}.

\keypoint{Simple-but-effective baselines} In competition with the plethora of sophisticated few-shot learners \cite{wang2020fslSurvey,hospedales2020meta} such as those mentioned above, a number of recent studies have advocated  strong baselines that perform comparably well while being simpler. These are often based on a transfer learning \cite{yosinski2014howTransferable} pipeline. They apply a conventional deep learner on the source data, before adapting to the few-shot target data by training a simple linear \cite{chen2019closerLook,tian2020rethinking,mangla2020charting} or centroid \cite{wang2019simpleshot} classifier on the fixed representation, or fine-tuning the feature backbone as well \cite{dhillon2020baselineFSL}. 
These methods mostly use standardized FSL source datasets (such as miniImageNet) and architectures (such as ResNet-12 and WRN-10-28) to enable direct comparisons of the advocated simple baselines to sophisticated learners. In contrast, we specifically aim to explore how far practical FSL performance can be pushed by exploiting other available pre-training datasets and architectures.

A few studies have evaluated FSL on a larger scale using datasets such as ImageNet1K  \cite{chen2021metaBaseline} or ImageNet21K \cite{dhillon2020baselineFSL}. However by changing both the source and target sets, this does not make it clear how choice/scale of source data impacts a given target problem -- the question that we answer here. Others have explored the impact of conventional  pre-training prior to meta-learning \cite{chen2021metaBaseline} or as a regularizer during meta-learning \cite{gidaris2019boosting} -- but without exploiting extra data. 

\keypoint{Bigger data and architectures} The impact of source datasets is widely studied in standard supervised \cite{sun2017unreasonable} and self-supervised \cite{goyal2019scaling,bommasani2021foundation} learning in vision, and in pattern recognition applications outside of vision \cite{devlin2019bert,brown2020languageGPT3,baevski2020wav2vec2nips,bommasani2021foundation}. However, it is not widely evaluated in FSL, which is a surprising omission, since as we shall see it may well be the easiest way to improve practical FSL performance. Similarly, existing FSL methods are almost exclusively based on a few less common architectures (e.g., Conv-4-64 and ResNet-12), which maybe due to the very first experimental setup on small datasets like Omniglot \cite{vinyals2016matching,finn2017model}.
Transformers have seen limited use in FSL, mainly for metric learning \cite{doersch2020crosstransformers}, but not for feature extraction. We explore how recent transformer feature extractors can be trained and applied to FSL, especially when combined with a foundation model \cite{bommasani2021foundation} pre-trained on larger source datasets.

\keypoint{Self-supervised \& few-shot} Our pipeline extends the typical unsupervised pre-train $\to$ supervised fine-tune workflow of the self-supervised research community \cite{ericsson2022ssrlSurvey,jing2021ssrlSurvey}, which has recently demonstrated strong performance for low-shot supervised learning  \cite{chen2020bigSimCLRV2,caron2021emergingDINO,ericsson2020well}. However, there has been limited direct comparison of self-supervised (SSL) and FSL community methods for data efficient learning due to different typical evaluation practices and benchmarks. For example, many SSL evaluations perform unsupervised representation learning on  ImageNet, before performing few-shot supervised learning within ImageNet \cite{chen2020bigSimCLRV2,caron2021emergingDINO}, which violates usual FSL community requirement of disjoint source and target data. One contribution of this paper is to provide a degree of comparison between and combination of the SSL and FSL approaches. For example, our MetaDataset, CDFSL and teaser Figure \ref{fig:teaser} results, use disjoint source and target data but benefit from external self-supervised pre-training.

\keypoint{Cross-domain few-shot} A FSL variant of particular practical interest is cross-domain few-shot \cite{guo2020broader}, where the source/meta-train dataset is significantly different to the target/meta-test dataset. This is more challenging than the standard within-domain setting, but more practically relevant. This is because in many scenarios where FSL is of interest such as medical or earth observation imaging \cite{guo2020broader}, the target data for FSL is significantly different to available source data (such as (mini-)ImageNet \cite{deng2009imagenet}). Major benchmarks of this type are CDFSL \cite{guo2020broader} and meta-dataset \cite{triantafillou2019meta}.

\section{A Simple Pipeline for FSL}\label{sec:method}

\keypoint{Problem formulation}
Few-shot learning (FSL) aims to learn a model with only a few annotated examples. %
One widely adopted formulation for FSL was introduced by Vinyals et al. \cite{vinyals2016matching} from a meta-learning perspective, where the assumption is that one should learn to solve new few-shot tasks based on previously seen  experience of many similar few-shot tasks. Therefore, the FSL problem is usually organized in two phases: \emph{meta-training} a few-shot learner on a distribution of training tasks and \emph{meta-testing} the resulting learner by evaluating it on novel few-shot tasks. Within each phase, data arrives in an episodic fashion, where the ``train-set'' and ``test-set'' of each task are called \emph{support set} and \emph{query set} respectively to avoid terminology confusion. In the case of classification, the difficulty level of an episode is described as \emph{K-way-N-shot}, which corresponds to learning a classifier for $K$ classes given $N$ examples per class in the support set. It is common to learn one model for each difficulty level, but a more realistic setting \cite{triantafillou2019meta} is to learn a global model for various K's and N's. This is sometimes called \emph{various-way-various-shot}, and we address this more practical setting here. This is also a reason to prefer simple pipelines over sophisticated meta-learners that may not be easily  extended to the various-way-various-shot setting.  

A different approach to small-data learning appears in the transfer learning \cite{bozinovski2020reminder,yosinski2014howTransferable} 
and self-supervision \cite{chen2020simpleCLR,bommasani2021foundation} literature. In this case one pre-trains a model using some large source data, and then re-purposes it for the sparse data target task of interest. The pre-training step aims to reduce the sample complexity of learning the target problem in the adaptation step. 

Although typically studied separately, both families of approach provide mechanisms for knowledge transfer from source data to the target few-shot problem of interest. Towards  the goal of high performance few-shot learning, we combine both pre-training (typically on auxiliary unlabeled data, which is freely and ubiquitously available) and meta-learning (episodic training with labels) together in a simple sequential pipeline using a single feature extractor backbone.
Our pipeline consists of three phases: 1) \textbf{pre-training} the feature backbone on unlabeled external data using self-supervised loss, 2) \textbf{meta-training} the feature backbone on labeled simulated few-shot tasks using ProtoNet \cite{snell2017prototypical} loss, and 3) deploying the feature backbone on novel few-shot tasks with optional  \textbf{fine-tuning} on the augmented support set of each task. A schematic of our pipeline is shown in Figure \ref{fig:overview}, which we call \ourName{} (i.e., the pipeline \ourNameLong{}). We next outline how the feature backbone is updated in different stages. 

\subsection{Pre-training of backbone}
We consider the feature backbones of ResNet \cite{he2016deep} or ViT \cite{dosovitskiy2021imageTransformerVIT}, to provide the foundation models in our pipeline. There are then several well-established self-supervised learning algorithms for the pre-training step: DINO \cite{caron2021emergingDINO} uses ImageNet1K and exploits the consistency in prediction between a large crop and multiple local crops of the same image, where a large crop is highly likely to overlap with a foreground object in the case of ImageNet images; BEiT \cite{bao2021beit} amounts to solving a masked image reconstruction task on the ImageNet-21K dataset in line with the original BERT pre-training \cite{devlin2019bert} for text data; and CLIP \cite{radford2021learning} leverages image captions in the YFCC100m dataset to align image and caption representations in a common feature space. 
For more flexible architectures like ViT \cite{dosovitskiy2021imageTransformerVIT}, pre-training on external data is important, as they are hard to train on common small-sized FSL benchmarks (Figure~\ref{fig:teaser} and Table~\ref{tab:q1q2}).

\subsection{Meta-training with ProtoNet}

As the goal is to build a simple pipeline, we consider the prototypical network (ProtoNet) \cite{snell2017prototypical}, which constructs class centroids dynamically for each episode and then performs nearest centroid classification. Specifically, ProtoNet only requires a feature backbone $f$ to map data points to a $m$-dimensional feature space: $f \colon \mathcal{X} \rightarrow \mathbb{R}^m$, and the probability of a query image $x$ belonging to class $k$ is given by
\begin{align}
    p(y=k | x) = \frac{\exp\big( -d(f(x), c_k) \big) }{\sum_{k'} \exp\big( -d(f(x), c_{k'}) \big)},
    \label{eq:proto}
\end{align}
where %
$d$ is implemented by a cosine distance in our work as opposed to the commonly chosen Euclidean distance and $c_k$ is the prototype of class $k$, defined as $c_k = \frac{1}{N_k} \sum_{i: y_i = k} f(x_i)$ and $N_k = \sum_{i: y_i = k} 1$ on the support set. %
Note that the prototypes can be computed regardless of the value of $k$. This enables ProtoNet to be trained and deployed under various-way-various-shot setting.

\subsection{Meta-testing with fine-tuning}\label{sec:method:fine-tune}

To be consistent with meta-training, by default, we deploy the meta-trained ProtoNet directly on all novel tasks. However, if the a novel task is drawn from an unseen domain, the learned feature representation may fail to generalize due to a substantial shift in the data distribution. To this end, we propose to fine-tune the feature backbone by a few gradient steps with the assistance of data augmentation. 
The details are summarized as PyTorch pseudo code in Algorithm \ref{algo:ft}. 

Our fine-tuning algorithm is similar to that of \cite{guo2020broader,li2021improvingFSL} who fine-tune the model weights using the support set since this is the only accessible labeled data at meta-test time. We exploit the support set slightly differently: we use data augmentation to create a pseudo query set derived from the support set; 
as such, we do not need to compute prototypes using the support set and then again apply the prototypes on the same support set using eq.~\eqref{eq:proto}. 
Besides, we simply update the entire backbone rather than exploring partial model adaptation.

\keypoint{Learning rate selection} 
We observe that the fine-tuning performance is relatively sensitive to the choice of learning rate (see supplemental material for more analysis). However, existing few-shot learning problem formulation does not offer a validation set for each task to choose the best learning rate for fine-tuning. Previous work \cite{guo2020broader,li2021improvingFSL} choose a learning rate a priori and fix it for every task. This strategy requires a good understanding of the backbone architecture but still leads to sub-optimal performance in general. 
Given a task with very few labeled images (i.e. the support set), it is almost unlikely to identify which learning rate yields good generalization for unlabeled images (i.e. the query set). The good news is that we find empirically the best learning rate is relatively stable across tasks within the same domain. 
To this end we propose to sample $N=5$ extra tasks from each domain and automate domain-wise learning rate search within a reasonable range (e.g., $\{0.01, 0.001, 0.0001, 0\}$). The best learning rate is then used for every task within the domain. This additional step amounts to preparing a few labeled images per domain to create a validation set, which makes sense in practice as we can easily organize tasks by domains and identify domain for individual tasks to look up the corresponding learning rate once searched.

\begin{algorithm}[tb]
   \caption{PyTorch pseudo code for fine-tuning}
   \label{algo:ft}
    \definecolor{codeblue}{rgb}{0.25,0.5,0.5}
    \lstset{
      basicstyle=\fontsize{7.2pt}{7.2pt}\ttfamily\bfseries,
      commentstyle=\fontsize{7.2pt}{7.2pt}\color{codeblue},
      keywordstyle=\fontsize{7.2pt}{7.2pt},
    }
\begin{lstlisting}[language=python]
# Inputs: a task including supp_x, supp_y, query_x
# backbone_state: meta-trained backbone weights
# optimizer: Adam optimizer 
# Outputs: logits

backbone = create_model_from_checkpoint(backbone_state)

def single_step(z):
    supp_f = backbone(supp_x) 
    proto = compute_prototypes(supp_f, supp_y)
    f = backbone(z) 
    logits = f.norm() @ proto.norm().T # cos similarity
    loss = cross_entropy_loss(logits, supp_y)
    return logits, loss

# fine-tuning loop    
for i in range(num_steps): 
    aug_supp_x = rand_data_augment(supp_x) 
    _, loss = single_step(aug_supp_x)
    loss.backward() # back-prop
    optimizer.step() # gradient descent
    
logits, _ = single_step(query_x) # classification 
\end{lstlisting}
\end{algorithm}

\section{Experiments}

\keypoint{Meta-training datasets}
We use standard benchmarks to evaluate our proposed pipeline. %
\textbf{miniImageNet} \cite{vinyals2016matching} contains 100 classes from ImageNet-1k, which is then split into 64 training, 16 validation  and 20 testing classes; each image is downsampled to 84$\times$84. 
\textbf{CIFAR-FS} ~\cite{bertinetto2018closedFormMeta}
 is created by dividing the original CIFAR-100 into
64 training, 16 validation  and 20 testing classes. The images are of size 32$\times$32.
\textbf{Meta-Dataset} \cite{triantafillou2019meta} subsumes 10 public image datasets of a diverse range of domains: ImageNet-1k, Omniglot, FGVC-Aircraft, CUB-200-2011, Describable Textures, QuickDraw, FGVCx Fungi, VGG Flower, Traffic Signs and MSCOCO. 
Each dataset has train/val/test splits. 
We follow the two training protocols proposed by \cite{triantafillou2019meta} and \cite{doersch2020crosstransformers} respectively. 
For the former, the train/val splits of the first 8 datasets (in-domain) are used for meta-training and validation, and the test splits of all datasets are used for meta-testing. The latter considers only ImageNet-1k's train-split for meta-training, and the other settings remain the same. 
For more details on Meta-Dataset %
we refer the readers to Appendix.3 of \cite{triantafillou2019meta}.

\keypoint{Evaluation} 
For evaluating few-shot classification performance, we simulate 600 episodes/tasks from the test-split for each dataset of interest. 
The evaluation metric is the average classification accuracy over tasks.
For miniImageNet and CIFAR-FS, the convention is to evaluate 5-way-1-shot (5w1s) and 5-way-5-shot episodes, and the size of the query set for each episode is fixed to $15 \times 5$. For Meta-Dataset, the number of ways, shots and query images are sampled uniformly at random with respect to the dataset specifications, except for ImageNet-1k and Omniglot (they have specific sampling strategies according to the hierarchy of classes). In addition, we evaluate the (5w5s) meta-trained model from miniImageNet for a cross-domain evaluation (\textbf{CDFSL}) \cite{guo2020broader}, where 4 out-of-domain datasets are considered, and the results are reported under 5-way-5/20/50-shot settings.

\keypoint{Training details}
To avoid over-engineering training for different datasets and  architectures, we adopt a common training strategy for meta-training the backbone from pre-trained model checkpoints (for both ResNet and ViT). This may lead to sub-optimal results for some cases, but it simplifies comparison. Specifically, we train the backbone for $100$ epochs, where each epoch consists of $2000$ episodes/tasks. We use a warm-up plus cosine annealing learning rate schedule: the learning rate starts from $10^{-6}$, increases to $5 \times 10^{-5}$ in $5$ epochs and then gradually decreases to $10^{-6}$ with a cosine annealing. We use the validation set to decide when to early stop,  
and turn off strong regularization and data augmentation techniques for simplicity.

\subsection{Analysis} 
We now use the pipeline outlined in Sec~\ref{sec:method} to answer a series of questions about few-shot learner pipeline design. Notably, \emph{\circled{1} How does pre-training regime affect FSL?} \emph{\circled{2} Can contemporary architectures such as ViT be adapted to FSL?} 
{\emph{\circled{3} How to exploit fine-tuning in meta-testing?}}

\begin{table}[t]
    \centering
    \resizebox{1.0\columnwidth}{!}{
    \begin{tabular}{llllccc}
    \toprule
       \multicolumn{4}{c}{Training Configuration} & \multicolumn{3}{c}{Benchmark Results}\\
    \hline
       ID & Arch & Pre Train & MetaTr & MD & miniIN & CIFAR\\
    \hline
        0&ViT-small & DINO (IN1K) & - & 67.4 & 97.0 & 79.8 \\
        1&ViT-small & DeiT (IN1K) & - & 67.5 & 98.8 & 84.6\\
        2&ResNet50 & DINO (IN1K) & - & 63.8 & 91.5 & 76.1 \\
        3&ResNet50 & Sup. (IN1K) & - & 62.4 & 96.4 & 82.3 \\
        \hline
        4&ViT-small & DINO (IN1K) & PN  & 78.4 & 98.0 & 92.5 \\
        5&ViT-small & DEIT (IN1K) & PN & 79.3 & 99.4 & 93.6 \\
        6&ViT-small & - & PN & 52.8 & 49.1 & 59.8 \\
        7&ResNet50 & DINO (IN1K) & PN & 72.4 & 92.0 & 84.0\\
        8&ResNet50 & Sup. (IN1K) & PN & 70.2 & 97.4 & 87.6 \\
        9&ResNet50 & - & PN & 62.9 & 72.2 & 68.4 \\
        \hline
        10&ResNet18 & - & PN & 63.3 %
         & \cut{61.1} %
        73.7 %
        & {70.2} \\
        \hline
        11 & ViT-base & DINO (IN1K) & PN & 79.2 & 98.4 & 92.2 \\
        12 & ViT-base & CLIP (YFCC) & PN & 80.0 & 98.1 & 93.2 \\
        13 & ViT-base & Sup (IN21K) & PN & 81.4 & 99.2 & 96.7 \\
        14 & ViT-base & BEIT (IN21K) & PN & 82.8 & 99.0 & 97.5 \\
        15 & ResNet50 & CLIP (YFCC) & PN & 75.0 & 92.2 & 82.6 \\
    \bottomrule
    \end{tabular}}
    \caption{The impact of architecture and pre-training algorithm (dataset) on downstream few-shot learning performance on Meta-Dataset (MD), miniImageNet (miniIN) and CIFAR-FS.  Meta-Dataset results are averaged over all target datasets while minIN and CIFAR results are 5-way-5-shot. 
    ProtoNet (PN)  nearest-centroid classifier is used throughout for few-shot learning on the support set during meta-test. MetaTr indicates the algorithm used for episodic learning on the corresponding benchmark.} 
    \label{tab:q1q2}
\end{table}

\subsubsection{Pre-training and architectures}\label{sec:q1q2}
We first evaluate the impact of pre-training regime (including algorithm and dataset), as well as neural architecture on FSL benchmarks Meta-Dataset \cite{triantafillou2019meta} (train on 8 datasets), miniImageNet \cite{vinyals2016matching}, and CIFAR-FS \cite{bertinetto2018closedFormMeta}. To clearly convey the configuration of each experiment, results in Table~\ref{tab:q1q2} are organized by architecture, pre-training algorithm (and dataset) and meta-training algorithm. 
We assume ProtoNet (nearest-centroid) classifier as the standard approach for meta-testing throughout, and compare either episodically trained ProtoNet or nothing as the meta-learning step between pre-training and meta-testing (column MetaTr).

\keypoint{\circled{1} How does pre-training regime affect FSL?} 
From the results in Table~\ref{tab:q1q2} we can draw the following conclusions: 
(i) Pre-training on ImageNet1K generally  provides a significant improvement across the board compared to the conventional pipeline used by prior work which does not make use of pre-training (compare model M9 with M7 and M8, etc\cut{; 6 with 4 and 5}). 
(ii) We are primarily interested in unsupervised pre-training, with supervised pre-training being included as an unfair upper bound. However, state of the art unsupervised pre-training with DINO performs close to supervised pre-training (compare M3 vs M2, etc). This is noteworthy, because while there is some semantic overlap between some of the source (ImageNet1K) and target (Meta-Dataset, miniImageNet, CIFAR) datasets considered here, good performance can be achieved without using source \emph{labels}, where there is no train-test label leakage\footnote{In the case of miniImageNet and Meta-Dataset, parts of ImageNet1K are used in both meta-train and meta-test splits. EG: since Meta-Dataset's ImageNet uses a 712/288 source/target class split, this means that for one of Meta-Dataset's 10 domains, there is some data (but not label) overlap between pre-train and meta-test for some foundation models. As discussed in Sec.~\ref{sec:related}, this overlap is ubiquitious in typical self-supervision evaluation pipelines \cite{caron2021emergingDINO,chen2020simpleCLR}. It is less common in FSL evaluation pipelines, but corresponds to making a semi-supervised or transductive assumption in terms of data access
as per \cite{ren2018metaSSL,li2019learning,huang2021lossConfSSLFSL,liu2019tpn}. Nevertheless, we do not think this is a significant factor in the strong results, as CLIP's YFCC does not have this overlap and performs similarly to the ImageNet1K based models.}. 
(iii) Given a strong pre-training regime such as DINO, simple nearest centroid classification based on pre-trained features performs well  (top block including M2, etc). {In particular, off-the-shelf features from a foundation model without dataset-specific meta-learning  perform favorably compared to conventional dataset-specific training of ProtoNet-ResNet18 (M2 vs M10), which is arguably the closest to industry standard in FSL.}  (iv) Nevertheless, dataset specific meta-learning does improve further (M7 vs M2, etc). Simple linear readout of a frozen foundation model \cite{ericsson2020well,chen2020bigSimCLRV2} is not competitive.

\keypoint{\circled{2} Can state of the art architectures such as ViT be adapted to FSL?} Using the results in Table~\ref{tab:q1q2}, we can also answer this question. In particular, while ViT does not train well on the smaller meta-train benchmarks (miniImageNet, CIFAR) compared to smaller architectures (see M6 vs M9, M10), it generally performs excellently when benefiting from large pre-training data (M6 vs M4). Overall ViT outperforms the industry standard ResNet18, as well as our ResNet50 baseline, across the board when benefitting from pre-training. We remark that our ResNet50 baseline also performs comparitively poorly without pre-training, especially on the smaller miniImageNet and CIFAR, suggesting that it is also too large to train well on the target datasets alone.

\keypoint{Other foundation models} 
Overall we can see that larger pre-training data sources, and recent architectures make a huge difference to downstream FSL performance on standard benchmarks. We also compared a selection of other foundation models \cite{bommasani2021foundation} in M11-15. We can see that (i) All the foundation models lead to substantial improvements on standard within-dataset training (M10,M9), (ii) The largest foundation models using, e.g., ViT-base and ImageNet21K or YFCC data source lead to strongest performance across the board, but do not outperform hugely the more economic DINO+ImageNet1K-based ViT-small (M4). For efficiency of pre-training and deployment, we take this to be our default model in the following section.

\begin{table}[t]
    \centering
    \resizebox{1.0\linewidth}{!}{
    \begin{tabular}{p{0.05\linewidth}p{0.17\linewidth}p{0.24\linewidth}p{0.18\linewidth}p{0.06\linewidth}p{0.06\linewidth}p{0.06\linewidth}p{0.06\linewidth}}
        \toprule
        \multicolumn{4}{c}{Train Config} & \multicolumn{4}{c}{Benchmark}\\
        \hline
        \multirow{2}{*}{ID} & \multirow{2}{*}{Arch} & \multirow{2}{*}{Pre~Train} & \multirow{2}{*}{MetaTr} & \multicolumn{2}{c}{miniIN} & \multicolumn{2}{c}{CIFAR}\\
        \multicolumn{4}{c}{} &  5/1 &  5/5 &  5/1 &  5/5\\
        \hline
        0&  ViT-small & DINO (IN1K) & - & 88.8 & 97.0 & 59.1 & 79.8\\
        1&ViT-small & DINO (IN1K) & ProtoNet & 93.1 & 98.0 & 81.1 & 92.5\\
        \hline
        2&ResNet18 & - & MetaQDA & 65.1 & 81.0 & - & -\\
        3&ViT-small & DINO (IN1K) & MetaQDA  & 92.0& 97.0 & 77.2 & 90.1\\
        \hline
        4&ResNet12 & - & MetaOptNet & 64.1 & 80.0 & 72.8 & 85.0\\
        5&ViT-small & DINO (IN1K) & MetaOptNet & 92.2 & 97.8 & 70.2 & 84.1\\
        \bottomrule
    \end{tabular}
    }
        \caption{Impact of architecture and pre-training on state-of-the-art few-shot learners: MetaQDA \cite{zhang2021metaQDA}, MetaOptNet \cite{lee2019meta}.}\label{tab:metaOpt}
\end{table}

\keypoint{\circled{1}+\circled{2} How does pre-training and architecture impact other Few-Shot Learners?} Our main experiments built upon ProtoNet as a widely used industry standard. We next explore how our pipeline impacts two few-shot learners that are more representative of recent state of the art, namely MetaOptNet \cite{lee2019meta} and MetaQDA \cite{zhang2021metaQDA}. From the results in Table~\ref{tab:metaOpt}, we can see that: (i) MetaQDA and MetaOptNet do improve on direct feature transfer (M5 and M3 vs M0) and on the simpler ResNet features they were initially evaluated with (M5 vs M4, M3 vs M2). But (ii) With the stronger features, they are outperformed by the simpler ProtoNet learner (M3 and M5 vs M1). This suggests previous conclusions about comparative meta-learner performance may need re-evaluating in this new regime of stronger features.

\keypoint{Few-shot learning v.s. self-supervised learning} Existing literature generally fails to directly compare algorithms from the few-shot learning community (such as ProtoNet,  \cite{snell2017prototypical}, MAML \cite{finn2017model}, MetaOptNet \cite{lee2019meta}, etc), with those from the self-supervised community (such as DINO \cite{caron2021emergingDINO}, SimCLR \cite{chen2020simpleCLR,chen2020bigSimCLRV2}, etc). This is partly because the popular evaluation protocol is different: For example 5-way-1-shot regime is popular the FSL community, vs 1\% labels ($\approx$ 1000-way-10-shot in the case of ImageNet) in the SSL community; network architectures differ ($\leq$ResNet18 vs $\geq$ResNet50 respectively); and image resolutions differ (84$\times$ vs full). Our results provide a taster of such a direct comparison. Overall they suggest that frozen self-supervised foundation models (using extra pre-training data) are  competitive out of the box compared to standard few-shot learners (using only meta-training data). However, more interestingly, combining these two paradigms as we have done, easily leads to state of the art performance on typical FSL metrics. 

\keypoint{Class overlap between pre-training and meta-testing} 
Although unsupervised pre-training does not utilize labels, it is very likely that some classes used by pre-training also appear in meta-testing. \emph{Does this class overlap go against the very definition of few-shot learning?} From a meta-learning point of view, the answer is yes. But we argue that class overlap is almost unavoidable unless a careful data split is simulated. For example, in the case of Meta-Dataset, the CUB dataset \cite{wah2011caltech}, the Aircraft dataset \cite{maji2013fine} and the COCO dataset \cite{lin2014microsoft} have a class overlap with ImageNet \cite{guo2016,doersch2020crosstransformers} but they are still used in meta-testing. As we consider more practical large-scale experiments, the class overlap issue becomes ubiquitous. We should worry about this issue if we were benchmarking a meta-learning algorithm, but for the nature of few-shot learning, benchmarking the capability of quickly constructing a classifier from very few labels is not hindered by class overlap. This is why self-supervised learning community is not bothered by this issue at all.  
It is worth mentioning that a similar setting called ``few-shot few-shot learning'' has been proposed by \cite{lifchitz2021few,zhang2021adargcn}, where they avoid overlap by either carefully picking up pre-training data from a different domain or crawling pre-training data of base categories from Internet. Alternatively, one may avoid overlap by using a different modality. We advocate meta-learning researchers to consider this controlled setting as a testing bed for incorporating powerful pre-trained feature backbones.

\begin{table}[t]
    \centering
    \resizebox{1.0\columnwidth}{!}{
    \begin{tabular}{llllllll}
    \toprule
       M& Arch & PreTr & MetaTr & MetaTe & Avg & Out-D \\
    \hline
        1&ViT-small & DINO & PN (IN) & PN & 68.38 & 67.68 \\
        2&ViT-small & DINO & PN (IN) & PN+FT(lr=0.01) & 76.05 & 76.54 \\
        3&ViT-small & DINO & PN (IN) & PN+FT(lr=0.001) & 74.47  & 74.51 \\
        4&ViT-small & DINO & PN (IN) & PN+FT(Tuned) & 77.53 & 77.85 \\
        \hline
        5&ViT-small & DINO & PN (MD) & PN & 78.43 & 55.71 \\
        6&ViT-small & DINO & PN (MD) & PN+FT(lr=0.01) & 76.09 & 73.26 \\
        7&ViT-small & DINO & PN (MD) & PN+FT(lr=0.001) & 74.64 & 69.97 \\
        8&ViT-small & DINO & PN (MD) & PN+FT(Tuned) & 83.13 & 75.72 \\
    \bottomrule
    \end{tabular}}
    \caption{Fine-tuning (FT) during meta-test on Meta-Dataset. The meta-train (MetaTr) setting indicates the source dataset as ImageNet only (IN) or full MetaDataset (MD). Results are the averages across all domains within meta-dataset (Avg), and just the out-of-distribution subset (Out-D). 
    }
\label{tab:q3q4}
\end{table}

\begin{figure}[ht]
\begin{center}
\includegraphics[width=0.95\linewidth]{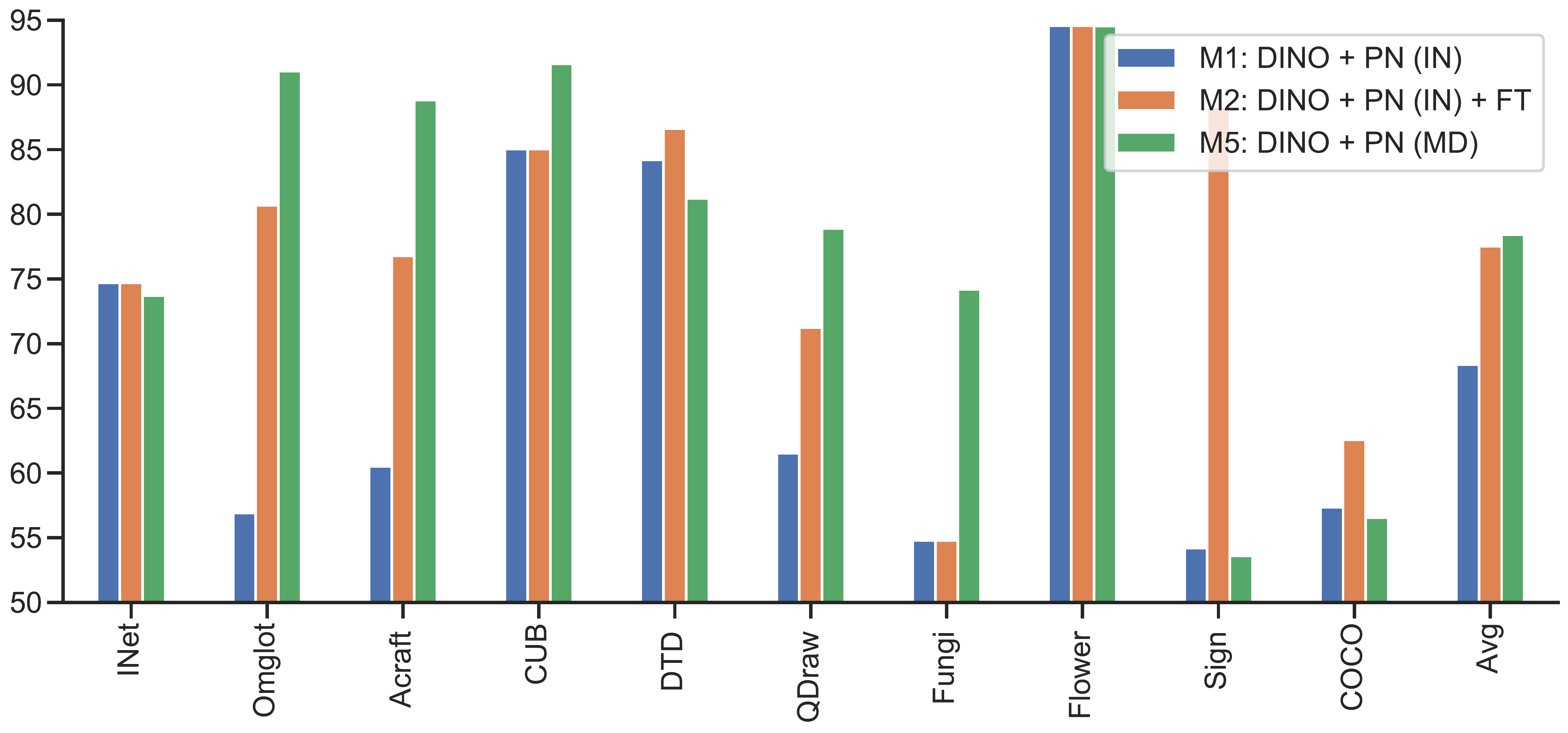}
\end{center}
\caption{The impact of fine-tuning during meta-test on Meta-Dataset. Held out datasets such as Signs and COCO benefit from fine-tuning; as do those very different from ImageNet such as omniglot and QuickDraw.
}
\label{fig:pt_md}
\end{figure} 

\subsubsection{Fine-tuning} 
The previous experiments used a fixed feature extractor together with ProtoNet for  meta-testing. We next investigate use of fine-tuning during meta-testing to further improve performance. We focus on the DINO pre-trained ViT models, based on their strong performance in Section~\ref{sec:q1q2}. %

\keypoint{\circled{3} How to best exploit fine-tuning for meta-testing?} To answer this question, we compare vanilla feature transfer as explored previously, with ProtoNet, and ProtoNet with episode-wise fine-tuning on the support set (ProtoNet+FT) as outlined in Section~\ref{sec:method:fine-tune}. We use Meta-Dataset including both conditions of treating ImageNet alone as the source, and joint meta-training on all of Meta-Dataset. From the results in Figure~\ref{fig:pt_md} and Table~\ref{tab:q3q4} we can draw the following conclusions: 
(i) Meta-training on the full Meta-Dataset improves on meta-training on ImageNet-training alone (M5 vs M1)\cut{, which in turn improves on direct feature transfer (M2 vs M1 vs M0)}. 
(ii) Fine-tuning during meta-test improves substantially in the out-of-distribution datasets, and especially in the case where meta-training is conducted on ImageNet, and then deployed across-domain to all the other Meta-Dataset tasks: See Out-D column and M2 vs M1 in Table~\ref{tab:q3q4}; blue vs orange bars in  Figure~\ref{fig:pt_md} for OmniGlot, QuickDraw, traffic signs, etc. However, for the condition where more Meta-Dataset domains are used for training and testing, fine-tuning has inconsistent impact across domains: While it is helpful for the remaining OOD datasets, it is not helpful overall (M5 vs M6 for Avg and Out-D). 
Overall feature backbone updates by fine-tuning are more helpful for domains unseen during meta-training, concurring with \cite{li2021improvingFSL,triantafillou2019meta}. On analysing the inconsistent impact of fine-tuning, we found this is due to difficulty in choosing an appropriate learning rate. Using any single learning rate throughout, as we did above (lr=0.01) is poorly tuned for some datasets. We therefore also explore our learning rate selection heuristic proposed in Section~\ref{sec:method:fine-tune}, 
and we see this leads to the best performance (M4 vs M2).

\begin{table}[t]
\centering
\resizebox{\linewidth}{!}{ %

\begin{tabular}{l|cc|l|l|l|l} 
\hline
\multirow{2}{*}{Method (Backbone)} & Ext. & Ext. & \multicolumn{2}{l|}{CIFAR-FS} & \multicolumn{2}{l}{MiniImageNet}  \\ 
\cline{4-7}
& dat. & lab.  & 5w1s  & 5w5s  & 5w1s  & 5w5s \\ 
\hline
\textbf{Inductive} &&&&&\\
ProtoNet (CNN-4-64) \cite{snell2017prototypical} && & 49.4 & 68.2 & 55.5 & 72.0 \\
Baseline++ (CNN-4-64) \cite{chen2019closerLook} & & & & & 48.2 & 66.4 \\
MetaOpt-SVM (ResNet12) \cite{lee2019meta}     &        &                               & 72.0    & 84.3       & 61.4  & 77.9 \\
Meta-Baseline (ResNet12) \cite{chen2021metaBaseline} & & &  &  & 68.6 & 83.7 \\ 
RS-FSL (ResNet12) \cite{afham2021rich} & & \ding{51}               &       &                   & 65.3  &  \\
\textbf{Transductive} &&&&&\\
Fine-tuning (WRN-28-10) \cite{dhillon2020baselineFSL} & & & 76.6 & 85.8 & 65.7 & 78.4 \\
SIB (WRN-28-10) \cite{Hu2020Empirical}     &&                               & 80.0    & 85.3            & 70.0    & 79.2 \\
PT-MAP (WRN-28-10) \cite{hu2021leveraging}            & &                    & \textbf{87.7} & 90.7      & 82.9 & 88.8 \\
CNAPS + FETI (ResNet18) \cite{bateni2020enhancing} & \ding{51}  & \ding{51}               &       &                   & 79.9  & 91.5 \\
\textbf{Self-supervised} &&&&&\\
ProtoNet (WRN-28-10) \cite{gidaris2019boosting} &&                               & 73.6  & 86.1              & 62.9  & 79.9 \\
ProtoNet (AMDIM ResNet) \cite{chen2021selfsupervised} & \ding{51} & & &  & 76.8 & 91.0 \\
EPNet + SSL (WRN-28-10) \cite{rodriguez2020embedding} & \ding{51} && & & 79.2 & 88.1 \\
\textbf{Semi-supervised} &&&&&\\
LST (ResNet12) \cite{li2019learning} & \ding{51} & & & & 70.1 & 78.7 \\
PLCM (ResNet12) \cite{huang2021lossConfSSLFSL} & \ding{51} & & 77.6 & 86.1 & 70.1 & 83.7 \\
\hline
P$>$M$>$F (IN1K, RN50)                              & \ding{51}    &                 & 73.7  & 84.0                  & 79.2  & 92.0 \\ 
P$>$M$>$F (IN1K, ViT-Small)                           & \ding{51}  &                   & 81.1  & \textbf{92.5}                  & 93.1  & 98.0 \\
P$>$M$>$F (IN1K, ViT-base)                               & \ding{51}   &                  & 84.3  & {92.2}         & \textbf{95.3}  & \textbf{98.4} \\
\hline
\end{tabular}

} %
\caption{
\textbf{miniImageNet \& CIFAR} -- Comparison with representative SOTA FSL algorithms. Methods using external data and/or labels are indicated. 
} %
\label{tab:single}
\end{table} %
\begin{table*}[ht]
\centering
\resizebox{\linewidth}{!}{ %
\begin{tabular}{l|llllllll|ll|l} 
\hline
\multirow{2}{*}{8 in-domain datasets} & \multicolumn{8}{c|}{In-domain}                                                                                & \multicolumn{2}{c|}{Out-of-domain} &         \\ 
\cline{2-12}
                                      & INet                           & Omglot & Acraft & CUB   & DTD   & QDraw & Fungi & Flower                     & Sign  & COCO                    & Avg     \\ 
\hline
ProtoNet \cite{triantafillou2019meta}  (RN18)                      & 67.01                          & 44.5   & 79.56  & 71.14 & 67.01 & 65.18 & 64.88 & 40.26                      & 86.85 & 46.48                   & 63.29  \\
CNAPs  \cite{requeima2020fast}  (RN18+Adapter)                           & 50.8                           & 91.7   & 83.7   & 73.6  & 59.5  & 74.7  & 50.2  & 88.9                       & 56.5  & 39.4                    & 66.90    \\ %
{SUR} \cite{dvornik2020selecting}   (RN18+Adapter)               & 57.2                           & 93.2   & \textbf{90.1}   & 82.3  & 73.5   & 81.9  & 67.9  & 88.4                       & 67.4  & 51.3                    & 75.32  \\ %
T-SCNAPs \cite{bateni2020enhancing}  (RN18+Adapter)    & 58.8                           & 93.9   & 84.1   & 76.8  & 69.0  & 78.6  & 48.8  & 91.6  & 76.1  & 48.7                    & 72.64   \\ %
URT \cite{liu2020universal}      (RN18+Adapter)         & 55.7                           & 94.4   & 85.8   & 76.3  & 71.8  & 82.5  & 63.5  & 88.2                       & 69.4  & 52.2                    & 73.98  \\ %
FLUTE \cite{triantafillou2021learning}   (RN18)    & 51.8                           & 93.2   & 87.2   & 79.2  & 68.8  & 79.5  & 58.1  & 91.6                       & 58.4  & 50.0                      & 71.78   \\
URL \cite{li2021universal}  (RN18+Adapter)      & 57.51                          & 94.51  & 88.59  & 80.54 & 76.17 & 81.94 & 68.75 & 92.11                      & 63.34 & 54.03                   & 75.75  \\
ITA \cite{li2021improvingFSL}      (RN18+Adapter)                         & 57.35                          & \textbf{94.96}  & 89.33  & 81.42 & 76.74 & \textbf{82.01} & 67.4  & 92.18                      & 83.55 & 55.75                   & 78.07  \\ 
\hline
P$>$M$>$F (DINO/IN1K, RN50)                 & 67.51                          & 85.91  & 80.3   & 81.67 & \textbf{87.08} & 72.84 & 60.03 & 94.69                      & 87.17 & 58.92                   & 77.61  \\
P$>$M$>$F (DINO/IN1K, ViT-small)               & 74.59                          & 91.79  & 88.33  & 91.02 & 86.61 & 79.23 & 74.2  & 94.12                      & 88.85 & 62.59                   & 83.13  \\
P$>$M$>$F (DINO/IN1K, ViT-base)               & \textbf{77.02}                          & 91.76  & {89.73}  & \textbf{92.94} & 86.94 & 80.2  & \textbf{78.28} & \textbf{95.79}                      & \textbf{89.86} & \textbf{64.97}                   & \textbf{84.75}  \\
\hline
\multirow{2}{*}{In-domain = ImageNet} & \multicolumn{1}{c|}{In-domain} & \multicolumn{9}{c|}{Out-of-domain}                                                                                &         \\ 
\cline{2-12}
                                      & \multicolumn{1}{c|}{INet}      & Omglot & Acraft & CUB   & DTD   & QDraw & Fungi & \multicolumn{1}{c}{Flower} & Sign  & COCO                    & Avg     \\ 
\hline
ProtoNet \cite{triantafillou2019meta} (RN18)                       & \multicolumn{1}{l|}{50.5}      & 59.98  & 53.1   & 68.79 & 66.56 & 48.96 & 39.71 & \multicolumn{1}{l}{85.27}  & 47.12 & 41                      & 56.10  \\
ALFA+FP-MAML \cite{baik2020metalearning} (RN12)   & \multicolumn{1}{l|}{52.8}      & 61.87  & 63.43  & 69.75 & 70.78 & 59.17 & 41.49 & \multicolumn{1}{l}{85.96}  & 60.78 & 48.11                   & 61.41  \\
BOHB \cite{saikia2020optimized}  (RN18)      & \multicolumn{1}{l|}{51.92}     & 67.57  & 54.12  & 70.69 & 68.34 & 50.33 & 41.38 & \multicolumn{1}{l}{87.34}  & 51.8  & 48.03                   & 59.15  \\
CTX \cite{doersch2020crosstransformers}  (RN34)  & \multicolumn{1}{l|}{62.76}     & 82.21  & 79.49  & 80.63 & 75.57 & 72.68 & 51.58 & \multicolumn{1}{l}{\textbf{95.34}}  & 82.65 & 59.9                    & 74.28  \\ 
\hline
P$>$M$>$F (DINO/IN1K, RN50)                 & \multicolumn{1}{l|}{67.08}     & 75.33  & 75.39  & 72.08 & 86.42 & 66.79 & 50.53 & \multicolumn{1}{l}{94.14}  & 86.54 & 58.2                    & 73.25   \\
P$>$M$>$F (DINO/IN1K, ViT-small)               & \multicolumn{1}{l|}{74.69}     & 80.68  & 76.78  & 85.04 & 86.63 & 71.25 & 54.78 & \multicolumn{1}{l}{94.57}  & 88.33 & 62.57                   & 77.53  \\
P$>$M$>$F (DINO/IN1K, ViT-base)               & \multicolumn{1}{l|}{\textbf{76.69}}     & \textbf{81.42}  & \textbf{80.33}  & \textbf{84.38} & \textbf{86.87} & \textbf{75.43} & \textbf{55.93} & \multicolumn{1}{l}{95.14}  & \textbf{89.68} & \textbf{65.01}                   & \textbf{79.09}  \\
\hline
\end{tabular}
} %
\caption{
\textbf{Meta-Dataset} -- Comparison with SOTA FSL algorithms.
}\label{tab:metadataset}
\end{table*} %
\begin{table*}[ht]
\centering
\resizebox{\linewidth}{!}{ %
\begin{tabular}{l|lll|lll|lll|lll} 
\hline
\multirow{2}{*}{}       & \multicolumn{3}{c|}{ChestX} & \multicolumn{3}{c|}{ISIC} & \multicolumn{3}{c|}{EuroSAT} & \multicolumn{3}{c}{CropDisease}  \\ 
\cline{2-13}
                        & 5w5s  & 5w20s & 5w50s       & 5w5s  & 5w20s & 5w50s     & 5w5s  & 5w20s & 5w50s        & 5w5s  & 5w20s & 5w50s            \\ 
\hline
ProtoNet  \cite{snell2017prototypical} (RN10)  & 24.05 & 28.21 & 29.32       & 39.57 & 49.50  & 51.99     & 73.29 & 82.27 & 80.48        & 79.72 & 88.15 & 90.81            \\
RelationNet  \cite{sung2018relationNet} (RN10)           & 22.96 & 26.63 & 28.45       & 39.41 & 41.77 & 49.32     & 61.31 & 74.43 & 74.91        & 68.99 & 80.45 & 85.08            \\
MetaOptNet  \cite{lee2019meta}  (RN10)     & 22.53 & 25.53 & 29.35       & 36.28 & 49.42 & 54.80      & 64.44 & 79.19 & 83.62        & 68.41 & 82.89 & 91.76            \\
Finetune   \cite{guo2020broader}  (RN10)             & 25.97 & 31.32 & 35.49       & 48.11 & 59.31 & 66.48     & 79.08 & 87.64 & 90.89        & 89.25 & 95.51 & 97.68            \\
CHEF  \cite{adler2021crossdomain}    (RN10)    & 24.72 & 29.71 & 31.25       & 41.26 & 54.30  & 60.86     & 74.15 & 83.31 & 86.55        & 86.87 & 94.78 & 96.77            \\
STARTUP \cite{phoo2021selftraining}    (RN10) & 26.94 & 33.19 & 36.91       & 47.22 & 58.63 & 64.16     & 82.29 & 89.26 & 91.99        & 93.02 & 97.51 & 98.45            \\
DeepCluster2 \cite{caron2020swav,ericsson2020well} (IN1K, RN50)& 26.51 & 31.51 & 34.17 & 40.73 & 49.91 & 53.65 & 88.39 & 92.02 & 93.07 & 93.63 & 96.63 & 97.04\\
\hline
P$>$M$>$F (DINO/IN1K, ResNet50) & 27.13 &	31.57 &	34.17 &	43.78 &	54.06 &	57.86 &	\textbf{89.18} &	\textbf{93.08} & \textbf{96.06} &	\textbf{95.06} &	97.25 &	97.77 \\
P$>$M$>$F (DINO/IN1K, ViT-small) & \textbf{27.27} & \textbf{35.33} & \textbf{41.39}       & \textbf{50.12} & \textbf{65.78} & \textbf{73.50}      & {85.98} & {91.32} & 95.40         & {92.96} & \textbf{98.12} & \textbf{99.24}           \\
\hline
\end{tabular}
} %
\caption{
\textbf{Broader study of cross-domain few-shot learning} -- Comparison with SOTA FSL algorithms.
}\label{tab:cdfsl}
\end{table*} 

\subsection{Results on standard benchmarks}

We call our pipeline \textbf{P>M>F}, which can be instantiated with any pre-training algorithm and backbone architectures, e.g., DINO > ProtoNet (PN) > Fine-tuning (FT). 
We next compare our pipeline with prior state of the art. \textbf{We emphasize that our results are not directly comparable to much prior SOTA in terms of architecture and use of external data.} We draw this comparison to see how simple changes (such as upgrading feature backbone to a modern network architecture and exploiting publicly available data for a large-scale pre-training) compare against 5 years of intensive research on FSL algorithms. The results for the single-domain cases, i.e., mini-ImageNet and CIFAR-FS, are summarized in \Table{single}, while the results for the cross-domain datasets, i.e., Meta-Dataset and Broader Study CDFSL, are shown in Table~\ref{tab:metadataset} and \ref{tab:cdfsl} respectively. From the results we can see that our framework outperforms much the state of the art in both within-domain and cross-domain conditions despite being significantly simpler than some sophisticated competitors. We remark that for the single source benchmarks in \Table{single}, a few competitors also used external data or ImageNet pre-training as indicated. Meanwhile our hybrid pipeline outperforms SOTA pure external self-supervision  \cite{caron2020swav,ericsson2020well} for CDFSL in Table~\ref{tab:cdfsl}. Our code is available at \url{https://github.com/hushell/pmf_cvpr22}.

\subsection{Discussion}
Taken together, the results  show that our simple pipeline of exploiting available pre-training data and  a modern architecture often outperforms sophisticated state of the art in few-shot learning.  This margin is increased using our proposed adaptive fine-tuning mechanism in the meta-test stage. Based on these observations we make recommendations both for practitioners and few-shot learning researchers. 

\noindent\textbf{Practitioners:} Increasing pre-training data size or simply using a foundation model \cite{caron2021emergingDINO,bommasani2021foundation} and upgrading  to modern architectures is likely to be more productive (and much easier to implement) than keeping up with and implementing state of the art few-shot learning algorithms. Fine-tuning is likely to be important if the target few-shot task of interest is less similar to the pre-training and meta-training data.

\noindent\textbf{FSL researchers:} Our results show that using external data and modern architectures is an easy and effective way to achieve strong FSL performance, and also that some SOTA meta-learners fail to provide expected improvements in this regime. While  external data violates definitions of the FSL problem that insist on a specific limited meta-train set, we should take this setting seriously to maintain practical relevance in the face of advancing self-supervision \cite{jing2021ssrlSurvey,ericsson2022ssrlSurvey,caron2021emergingDINO,radford2021learning}. In particular, we recommend a new evaluation setting for all the standard FSL benchmarks, where pre-train data and architecture are freely chosen and clearly reported. Few-shot meta-learning methods are then evaluated on their ability to improve on linear readout, fine-tuning, or our PMF baseline for the given external dataset and architecture.

\section{Conclusions}

We advanced few-shot learning from the perspective of pushing the limits of a simple  pre-train + ProtoNet pipeline in terms of dataset, architecture and fine-tuning strategy. We showed that source dataset, and neural architecture are dominant factors in FSL performance.
When there is a domain shift between training and testing, we showed that fine-tuning the feature backbone with data augmentation is also important. 
We verified that our simple pipelines achieve very competitive performance in four FSL benchmarks.

\keypoint{Limitations and future work} There are several limitations of our empirical study. We only scratched the surface of the impact of external data and correspondingly larger architectures on FSL. Our renewed focus on external data emphasizes the need for algorithms from the FSL community \cite{finn2017model,snell2017prototypical,lee2019meta} to be directly compared against algorithms from the self-supervised community \cite{chen2020simpleCLR,bommasani2021foundation}, or possibly synergistically combined, as we attempt here. The hybrid pipeline that we propose is obviously restricted to modalities where large external datasets already exist, and would require significant up-front investment in compute and energy cost where pre-trained foundation models do not already exist. Possible bias within foundation models is also a potential risk \cite{bommasani2021foundation}. Finally, while effective, our adaptive fine-tuning strategy, is rather computationally expensive at meta-test time, and may be unsupported on embedded platforms without backpropagation. Feed-forward representation adaptation methods \cite{requeima2020fast} may be important for future work.

\section*{Acknowledgement}
We thank the anonymous reviewers and meta-reviewers of CVPR2022 for their careful reading and thorough discussion of our manuscript. We also thank our colleagues at SAIC-Cambridge, especially Gabor Gyorkei, Taekwon Jang and Brais Martinez, for their help and support.  

{
    \small
    \bibliographystyle{ieee_fullname}
    \bibliography{macros,main}
}



\end{document}


\title{Pushing the Limits of Simple Pipelines for Few-Shot Learning: \\ Supplemental Material}
\author{}
\maketitle

In this supplemental material, we present: 

\begin{itemize}
\item[$\bullet$] In Section~\ref{sec:tab1}, we include additional results for Table 1 in the main paper. 
\item[$\bullet$] In Section~\ref{sec:tab4}, we include additional results for Table 1 and Table 4 in the main paper.
\item[$\bullet$] In Section~\ref{sec:ftlr}, we investigate the impact of the hyper-parameters for the fine-tuning phase. 
\item[$\bullet$] In Section~\ref{sec:tsne}, we show the T-SNE plots before and after ProtoNet meta-training.
\end{itemize}


\section{Additional results for Meta-Dataset} \label{sec:tab1}

In this section, we show a complete view of the results presented in Table 1 in the main paper, including the outcomes of different pre-training methods (see Table~\ref{tab:pt}), the outcomes of meta-training on ImageNet domain (see Table~\ref{tab:in}), and the outcomes of meta-training on eight pre-specified domains (see Table~\ref{tab:md}). 

As indicated in the main paper, our pipeline is named in a form of ``P > M > F (backbone)'', where ``P'', ``M'' and ``F'' are taken from the first letters of pre-training, meta-training and fine-tuning respectively. In this section, we only examine the pre-training and backbone architecture parts with meta-training fixed to ProtoNet. As an example, in Table~\ref{tab:in}, we use ``DINO > PN (ViT-small)'' to denote the pipeline that uses DINO pre-training, ProtoNet meta-training with backbone architecture being ViT-small.  

To clarify the shorten notations in Table~\ref{tab:pt}, Table~\ref{tab:in} and Table~\ref{tab:md}, we make a list here: 
\begin{itemize}
    \item DINO: self-distillation pre-training on ImageNet-1k dataset by \cite{caron2021emergingDINO}.
    \item BEiT: BERT pre-training on ImageNet-21k dataset by \cite{bao2021beit}.
    \item CLIP: Contrastive language-image pre-training on YFCC100M dataset by \cite{radford2021learning}. 
    \item Sup21k: Supervised pre-training on ImageNet-21k dataset.
    \item Sup1k: Supervised pre-training on ImageNet-1k dataset.
    \item BEiT + Sup21k: BERT unsupervised pre-training first on ImageNet-21k dataset and then using the labels of ImageNet-21k to fine-tune the model.
\end{itemize}

\begin{table}[h]
\centering
\centering
\resizebox{\linewidth}{!}{ 
\begin{tabular}{l|llllllllll|l} 
\hline
\hline
                         & INet  & Omglot & Acraft & CUB   & DTD   & QDraw & Fungi & Flower & Sign  & COCO  & Avg     \\ 
\hline
DINO (ViT-small)         & 73.48 & 54.33  & 62.17  & 85.37 & 83.67 & 60.59 & 56.26 & 94.45  & 53.7  & 54.58 & 67.86   \\
DINO (ViT-base)          & 74.85 & 59.44  & 55.36  & 80.08 & 84    & 59.61 & 56.65 & 94.84  & 51.81 & 57.1  & 67.374  \\
BEiT (ViT-base)          & 17.12 & 23.96  & 17.21  & 18.59 & 39.79 & 23.89 & 13.69 & 45.81  & 16.16 & 16.36 & 23.258  \\
CLIP (ViT-base)          & 60.66 & 62.12  & 54.08  & 80.26 & 76.51 & 62.90 & 30.76 & 68.43  & 47.33 & 41.95 & 58.5    \\
DINO (ResNet50)          & 64.13 & 52.51  & 57.02  & 62.63 & 84.5  & 60.78 & 50.41 & 92.18  & 58.27 & 55.43 & 63.786  \\
CLIP (ResNet50)          & 51.67 & 44.16  & 44.18  & 70.2  & 70.64 & 47.88 & 34.13 & 87.97  & 39.59 & 41.63 & 53.205  \\
Sup21k (ViT-base)        & 67.00    & 37.02  & 47.72  & 82.9  & 79.77 & 52.25 & 41.98 & 95.7   & 46.22 & 53.46 & 60.402  \\
BEiT + Sup21k (ViT-base) & 33.85 & 23.95  & 33.92  & 52.07 & 63.79 & 32.60 & 28.19 & 67.3   & 27.18 & 29.65 & 39.25   \\
Sup1k (ViT-base)         & 89.1  & 60.71  & 55.36  & 79.8  & 79.75 & 61.28 & 47.45 & 88.44  & 56.3  & 57.20 & 67.539  \\
Sup1k (ResNet50)         & 76.22 & 47.31  & 55.75  & 76.40 & 80.40 & 51.26 & 43.42 & 85.48  & 50.46 & 57.10 & 62.38   \\
\hline
\hline
\end{tabular}
} 
\caption{
\textbf{Pre-training results on Meta-Dataset} -- Comparison of different pre-training methods and backbone architectures.
}\label{tab:pt}
\end{table} 
\begin{table}[h]
\centering
\centering
\resizebox{\linewidth}{!}{ 

\begin{tabular}{ll|l|lllllllll|l} 
\hline
\hline
 &                                              & In-domain & \multicolumn{9}{c|}{Out-of-domain}                                       &         \\ 
\cline{3-12}
 &                                              & INet      & Omglot & Acraft & CUB   & DTD   & QDraw & Fungi & Flower & Sign  & COCO  & Avg     \\ 
\hline
\multicolumn{2}{l|}{DINO > PN (ViT-small)}         & 74.69     & 56.91  & 60.5   & 85.04 & 84.21 & 61.54 & 54.78 & 94.57  & 54.21 & 57.35 & 68.38   \\
\multicolumn{2}{l|}{DINO > PN (ViT-base)}          & 76.69     & 62.2   & 54.76  & 81.58 & 84.48 & 60.64 & 55.93 & 95.14  & 56.81 & 60.27 & 68.85   \\
\multicolumn{2}{l|}{CLIP > PN (ViT-base)}          & 76.03     & 59     & 65.75  & 90.2  & 83.08 & 65.45 & 53.2  & 96.35  & 58.65 & 61.2  & 70.891  \\
\multicolumn{2}{l|}{DINO > PN (ResNet50)}          & 67.08     & 49.21  & 58.46  & 72.08 & 85.01 & 59.2  & 50.53 & 89.91  & 55.44 & 53.94 & 64.086  \\
\multicolumn{2}{l|}{CLIP > PN (ResNet50)}          & 69.41     & 60.72  & 57.53  & 83.66 & 80.03 & 55.58 & 50.07 & 93.39  & 48.56 & 50.14 & 64.909  \\
\multicolumn{2}{l|}{Sup21k > PN (ViT-base)}        & 85.88     & 39.72  & 52.03  & 94.54 & 83.42 & 54.58 & 57.06 & 99.01  & 47.74 & 69.02 & 68.3    \\
\multicolumn{2}{l|}{BEiT+Sup21k > PN (ViT-base)}   & 84.39     & 60.54  & 74.04  & 95.66 & 86.14 & 65.24 & 64.25 & 99.19  & 63.02 & 69.91 & 76.238  \\
\multicolumn{2}{l|}{Sup1k > PN (ViT-base)}         & 90.48     & 62.96  & 54.89  & 78.88 & 80.02 & 61.81 & 45.52 & 88.56  & 55.61 & 59.12 & 67.785  \\
\hline
\hline
\end{tabular}

} 
\caption{
\textbf{Meta-training results on Meta-Dataset (ImageNet only)} -- Comparison of different pre-training methods and backbone architectures.
}\label{tab:in}
\end{table}
\begin{table}[h]
\centering
\centering
\resizebox{\linewidth}{!}{ 

\begin{tabular}{l|llllllll|ll|l} 
\hline
\hline
                         & \multicolumn{8}{c|}{In-domain}                                   & \multicolumn{2}{l|}{Out-of-domain} &         \\ 
\cline{2-11}
                         & INet  & Omglot & Acraft & CUB   & DTD   & QDraw & Fungi & Flower & Sign  & COCO                       & Avg     \\ 
\hline
DINO > PN (ViT-small)       & 73.54 & 91.79  & 88.33  & 91.02 & 81.64 & 79.23 & 74.2  & 94.12  & 54.37 & 57.04                      & 78.528  \\
DINO > PN (ViT-base)        & 73.55 & 91.54  & 89.73  & 92.94 & 81.52 & 80.2  & 78.28 & 94.53  & 53.65 & 59.13                      & 79.507  \\
CLIP > PN (ViT-base)        & 74.76 & 92.26  & 91.42  & 93.55 & 80.97 & 80.8  & 79.13 & 95.64  & 54.52 & 56.8                       & 79.985  \\
DINO > PN (ResNet50)        & 63.7  & 85.91  & 80.3   & 81.67 & 82.69 & 72.84 & 60.03 & 91.75  & 54.26 & 50.67                      & 72.382  \\
CLIP > PN (ResNet50)        & 64.86 & 92.09  & 89.19  & 89.17 & 71.67 & 78.71 & 76.15 & 91.25  & 51.1  & 45.88                      & 75.007  \\
Sup21k > PN (ViT-base)      & 84.86 & 85.71  & 83.77  & 95.89 & 85.1  & 78.47 & 74    & 99.17  & 59.86 & 67.57                      & 81.44   \\
BEiT+Sup21k > PN (ViT-base) & 81.96 & 94.19  & 91.62  & 93.76 & 81.3  & 83.48 & 81.76 & 98.84  & 58.83 & 61.81                      & 82.755  \\
Sup1k > PN (ViT-small)      & 83.87 & 91.22  & 87.9   & 89.2  & 78.11 & 78.7  & 70.33 & 94     & 56.24 & 57.16                      & 78.673  \\
Sup1k > PN (ViT-base)       & 89.75 & 93.48  & 91.15  & 92.48 & 78.52 & 80.65 & 75.97 & 95.78  & 53.47 & 55.89                      & 80.714  \\
Sup1k > PN (ResNet50)       & 68.04 & 86.17  & 80.72  & 80.48 & 71.65 & 70.78 & 59.58 & 84.33  & 50.06 & 50.29                      & 70.21   \\
None > PN (ViT-small)       & 37.25 & 74.14  & 45.25  & 49.66 & 61.49 & 70.24 & 43.23 & 72.03  & 39.33 & 35.43                      & 52.805  \\
None > PN (ResNet50)        & 40.74 & 90.67  & 80.67  & 68.88 & 62.4  & 75.96 & 55.72 & 75.37  & 43.11 & 35.49                      & 62.901  \\
\hline
\hline
\end{tabular}

} 
\caption{
\textbf{Meta-training results on Meta-Dataset} -- Comparison of different pre-training methods and backbone architectures.
}\label{tab:md}
\end{table}

\section{Additional results for miniImageNet and CIFAR-FS} \label{sec:tab4}

We also evaluate different pre-training methods and backbones on miniImageNet and CIFAR-FS, which is shown in Table~\ref{tab:m&c}. We do not include some of the results to the main paper because supervised pre-training on ImageNet is only useful to check the upper bound performance. 

\begin{table}[ht]
\centering

\begin{tabular}{l|ll|ll} 
\hline
\hline
                         & \multicolumn{2}{l|}{miniImageNet} & \multicolumn{2}{l}{CIFAR-FS}  \\ 
\hline
                         & 5w1s & 5w5s                        & 5w1s & 5w5s                   \\ 
\cline{2-5}
DINO > PN (ViT-small)       & 93.1 & 98.0                          & 81.1 & 92.5                   \\
DINO > PN (ViT-base)        & 95.3 & 98.4                        & 84.3 & 92.2                   \\
CLIP > PN (ViT-base)        & 93.1 & 98.1                        & 85.3 & 93.2                   \\
DINO > PN (ResNet50)        & 79.2 & 92.0                        & 73.7 & 84.0                   \\
CLIP > PN (ResNet50)        & 78.9 & 92.2                        & 71.4 & 82.6                   \\
Sup21k > PN (ViT-base)      & 97.2 & 99.2                        & 92.3 & 96.7                   \\
BEiT+Sup21k > PN (ViT-base) & 96.6 & 99                          & 93.8 & 97.5                   \\
Sup1k > PN (ViT-small)      & 97.7 & 99.4                        & 86.2 & 93.6                   \\
Sup1k > PN (ViT-base)       & 99.2 & 99.8                        & 88.2 & 94.3                   \\
Sup1k > PN (ResNet50)       & 91.7 & 97.4                        & 77   & 87.6                   \\
None > PN (ViT-small)       & 36.5 & 49.1                        & 45.9 & 59.8                   \\
None > PN (ResNet50)        & 46.1 & 60.3                        & 54.1 & 68.4                   \\
\hline
\hline
\end{tabular}

\caption{
\textbf{miniImageNet \& CIFAR-FS} -- Comparison of different pre-training methods and backbone architectures. 
} 
\label{tab:m&c}
\end{table}


\section{Ablation study on fine-tuning's hyper-parameters} \label{sec:ftlr}

\begin{figure*}[ht]
    \centering
	\begin{minipage}[t]{0.48\textwidth}  
		\centering 
        \includegraphics[width=\textwidth]{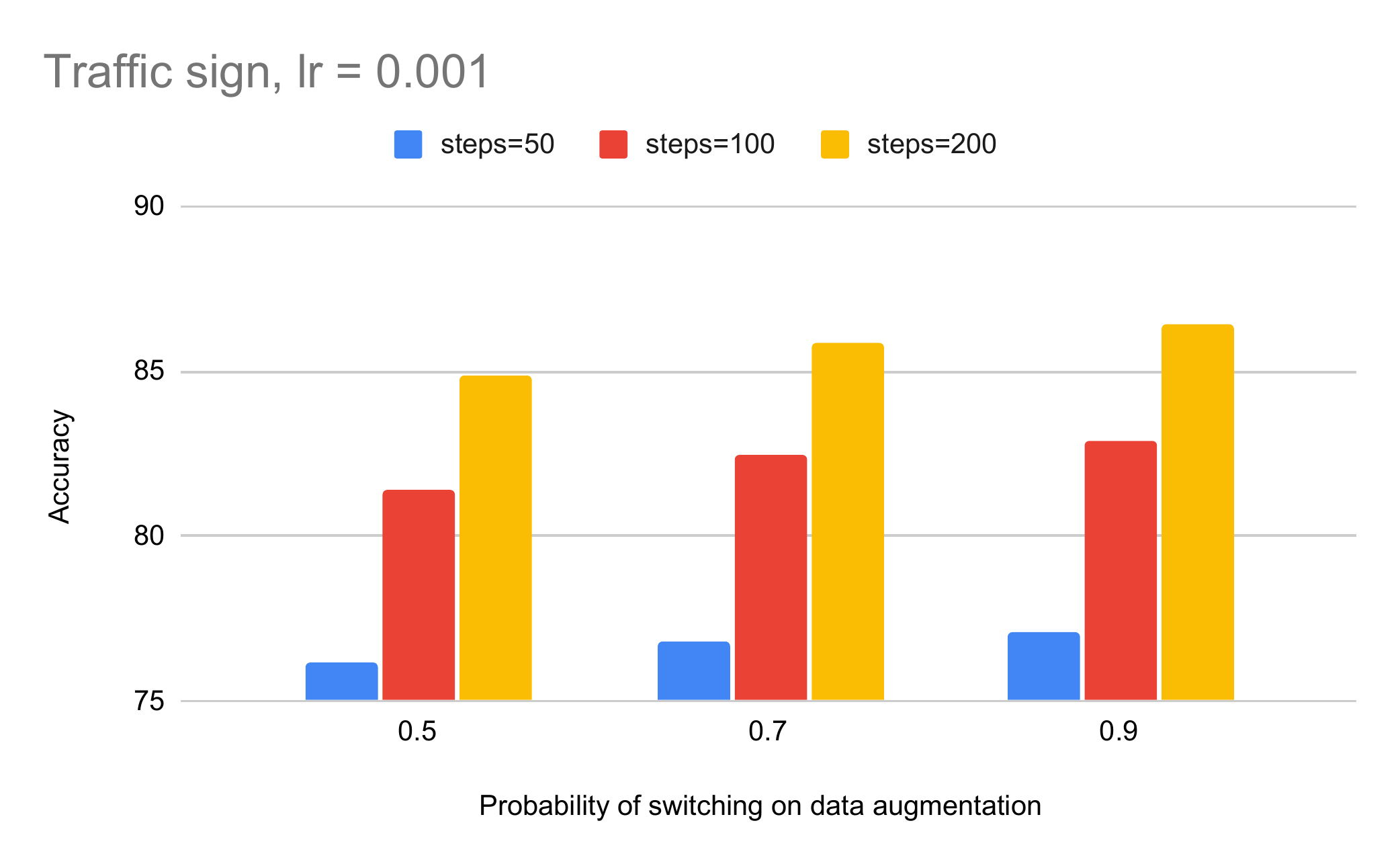}
	\end{minipage}  
	\begin{minipage}[t]{0.48\textwidth}  
		\centering  
        \includegraphics[width=\textwidth]{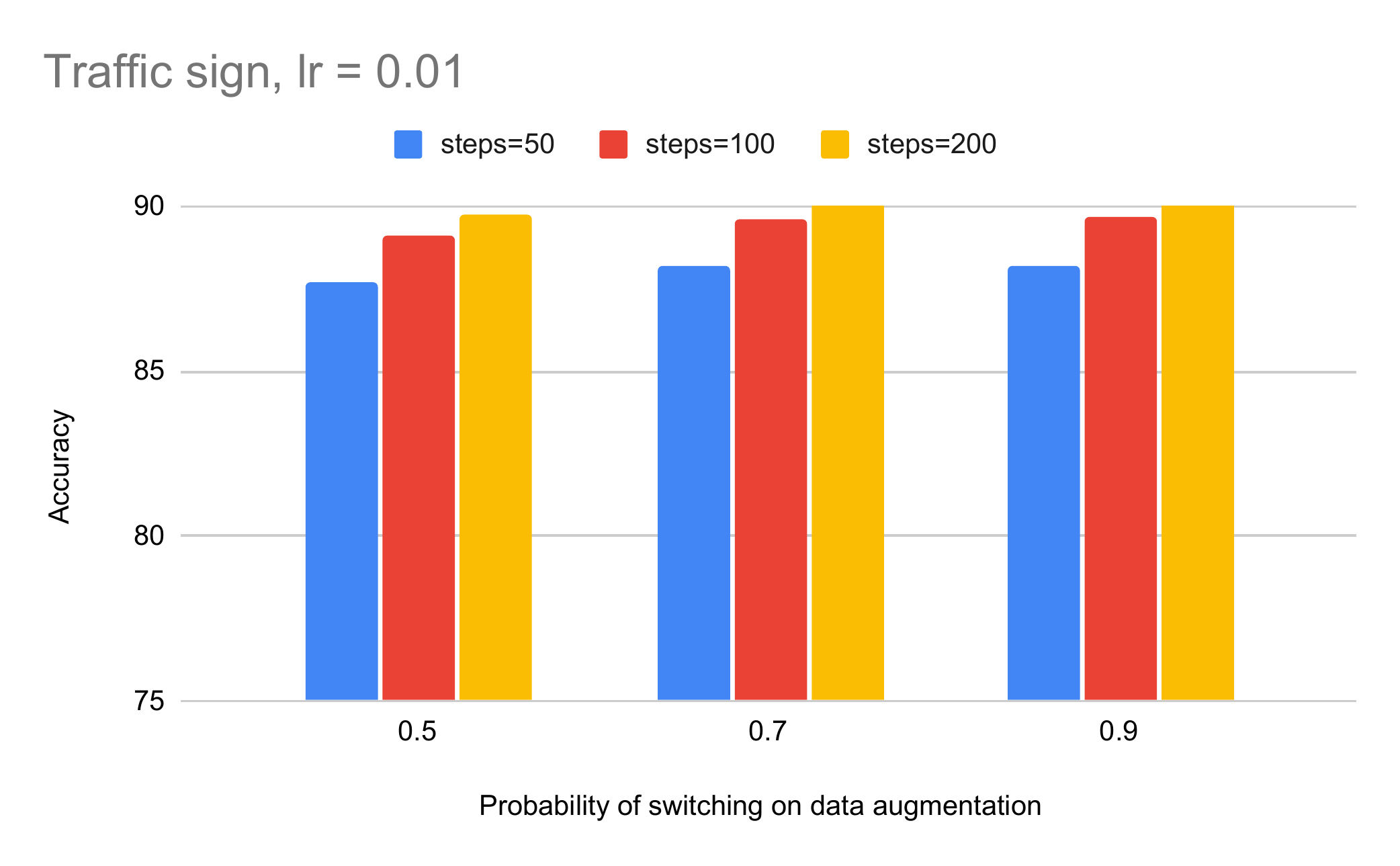} 
	\end{minipage} \\
	\begin{minipage}[t]{0.48\textwidth}  
		\centering  
        \includegraphics[width=\textwidth]{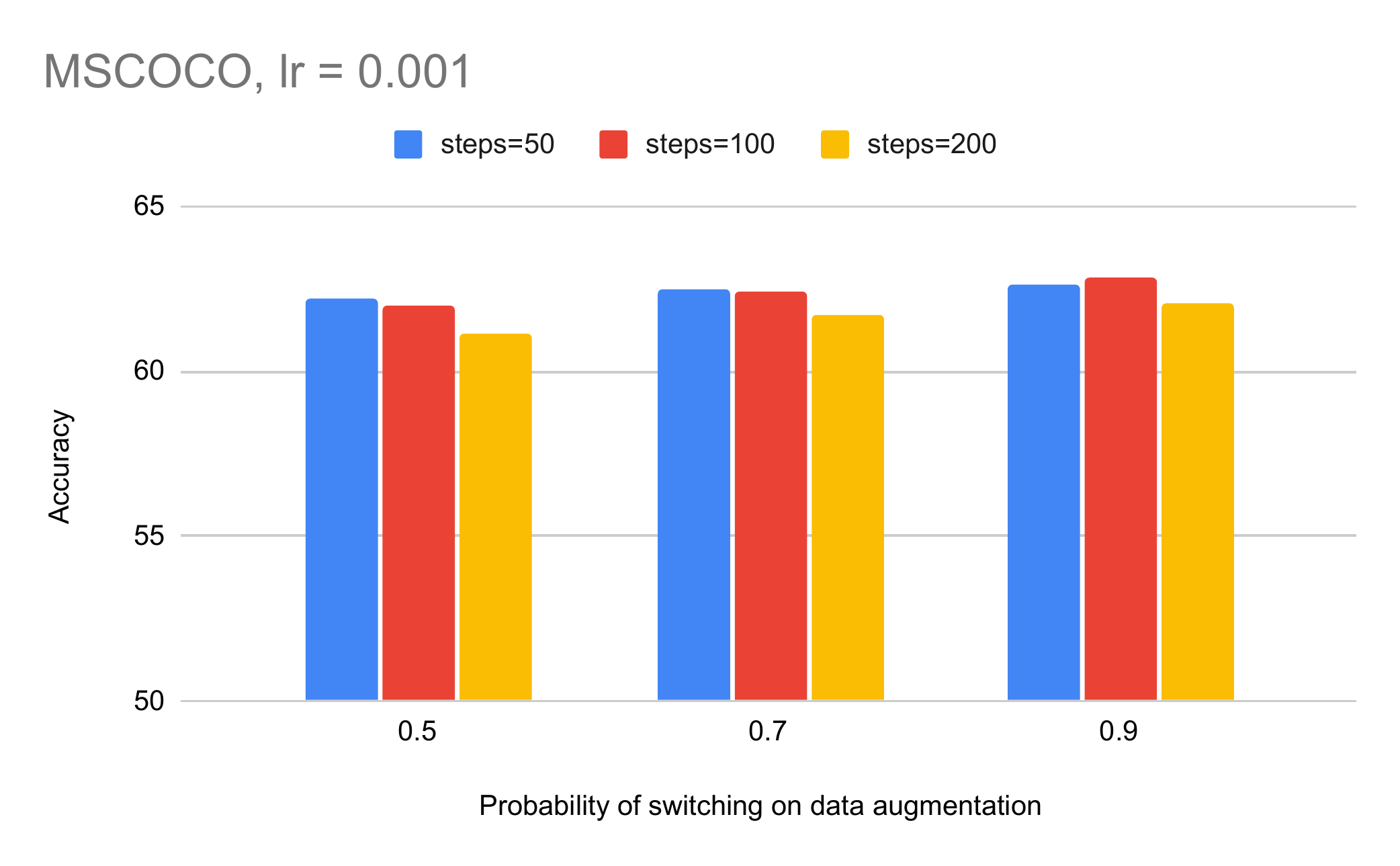} 
	\end{minipage} 
	\begin{minipage}[t]{0.48\textwidth}  
		\centering  
        \includegraphics[width=\textwidth]{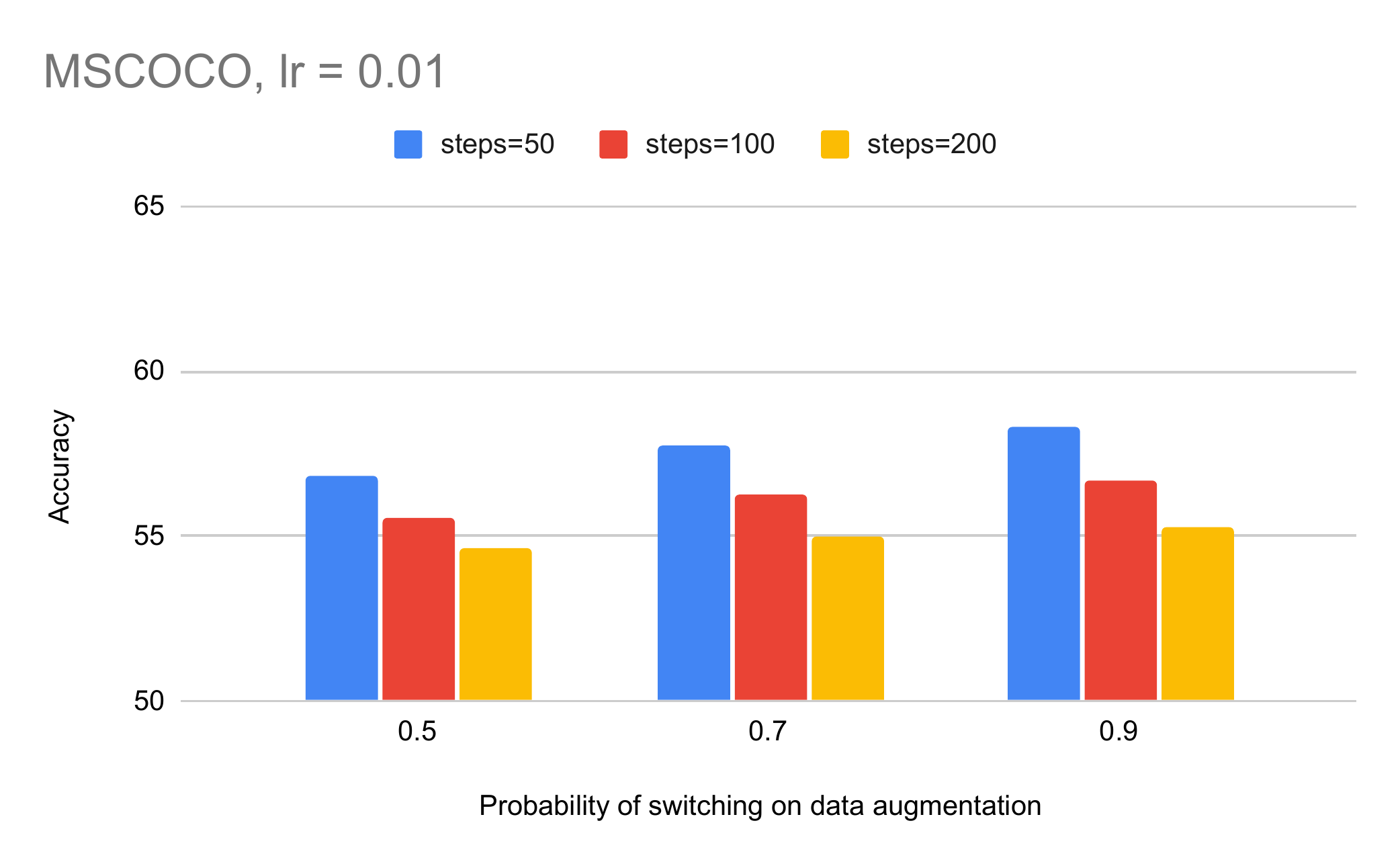}
	\end{minipage} 
	\caption{\textbf{Ablation study of fine-tuning's hyper-parameters} -- The experiments are done in the validation set of the traffic sign domain and the MSCOCO domain with learning rate fixed to either 0.001 or 0.01.}
	\label{fig:ft_ablation}
\end{figure*}

There are three hyper-parameters for the fine-tuning stage: the learning rate, the number of gradient descent steps and the probability of switching on data augmentation for the support set. We show in Figure~\ref{fig:ft_ablation} that the dominant hyper-parameter is the learning rate. From the results, we also see that the higher the probability of switching on data augmentation the better, while 50 gradient steps give relatively good performance with the right learning rate. Therefore, we fix the probability to 0.9 and let the numbers of steps to be 50 in the fine-tuning phase.

\section{T-SNE plots: before and after meta-training} \label{sec:tsne}

\begin{figure*}[t]
    \centering
    \includegraphics[width=\textwidth]{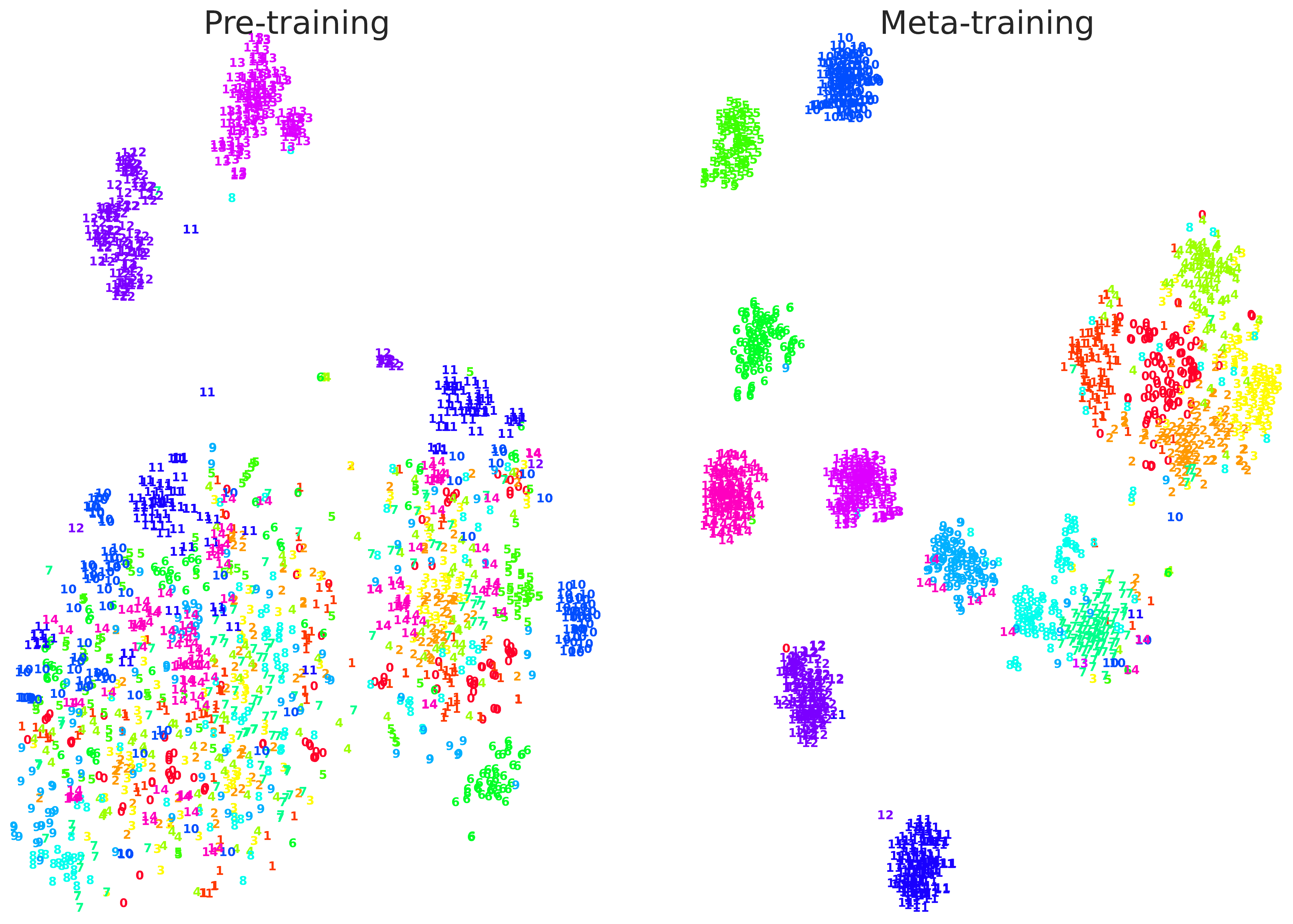}
    \caption{\textbf{Aircraft domain}}
	\label{fig:aircraft}
\end{figure*}

\begin{figure*}[t]
    \centering
    \includegraphics[width=\textwidth]{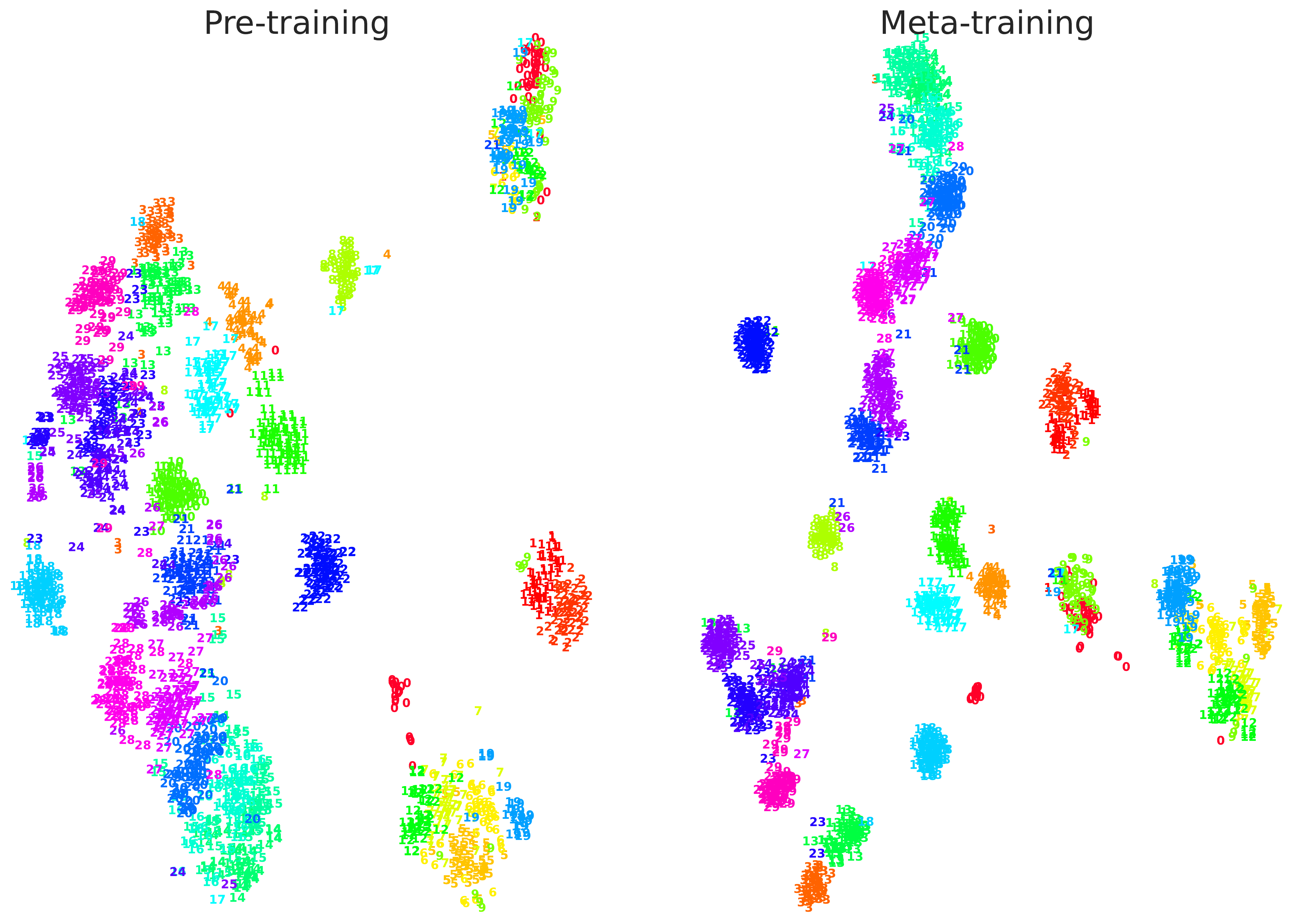}
    \caption{\textbf{CUB domain}}
	\label{fig:cub}
\end{figure*}

\begin{figure*}[t]
    \centering
    \includegraphics[width=\textwidth]{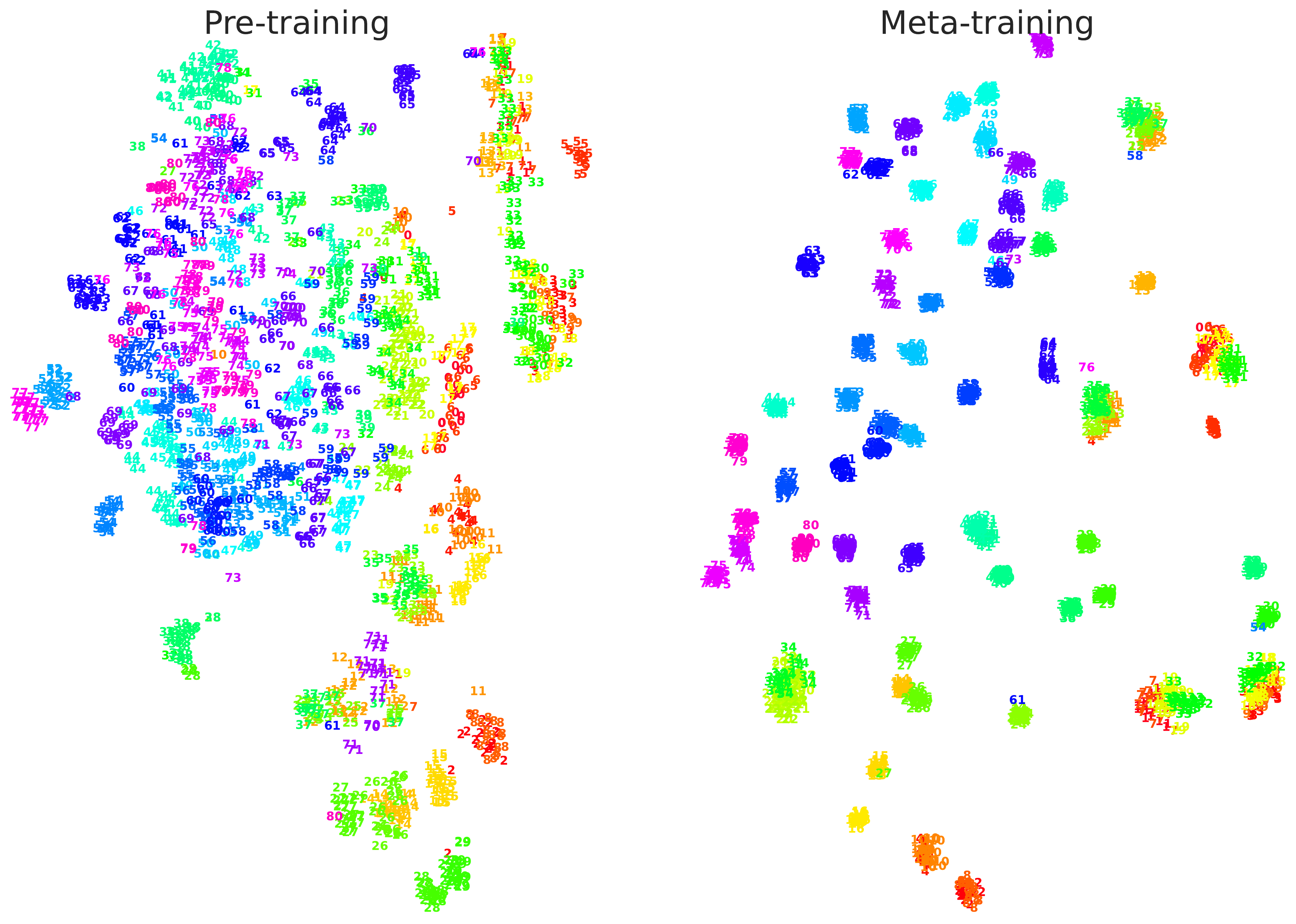}
    \caption{\textbf{Omniglot domain}}
	\label{fig:omniglot}
\end{figure*}

By using T-SNE visualization, We identify that the feature representation of DINO pre-training is already of high quality in multiple domains. Three examples are shown in Figure~\ref{fig:aircraft}, Figure~\ref{fig:cub} and Figure~\ref{fig:omniglot}. In general, many semantic clusters have already emerged, even though these domains where the clusters are sitting are not necessarily similar to ImageNet. This gives a very good initialization to ProtoNet so that it can refine the clusters to be much tighter. While the situation would be quite different if we were training the ProtoNet from scratch, which are confirmed by the no-pre-training results in Table~\ref{tab:md}. This can be explained in the sense of K-means clustering, where a good initialization is always desired.    

{
    \clearpage
    \small
    \bibliographystyle{ieee_fullname}
    \bibliography{macros,main}
}